%% file: starai-2018.tex
\documentclass[conference]{IEEEtran}
\usepackage{fancyhdr}
\ifCLASSINFOpdf
\else
\fi
\usepackage[inference]{semantic}

\usepackage{amsmath}
\usepackage{multirow}

\input{preamble-stuff}

\begin{document}
	\IEEEoverridecommandlockouts
	
	\IEEEpubid{\makebox[\columnwidth]{978-1-4799-7560-0/15/\$31 \copyright 2015 IEEE \hfill} \hspace{\columnsep}\makebox[\columnwidth]{ }}
%
% paper title
% can use linebreaks \\ within to get better formatting as desired
%\title{Outlier Detection For Object-Relational Data}
\title{Model-based Exception Mining for Object-Relational Data}

% author names and affiliations
% use a multiple column layout for up to three different
% affiliations
\author{\IEEEauthorblockN{Fatemeh Riahi}
\IEEEauthorblockA{School of Computing Science\\
Simon Fraser University\\
Burnaby, Canada\\
sriahi@sfu.ca}
\and
\IEEEauthorblockN{Oliver Schulte}
\IEEEauthorblockA{School of Computing Science\\
Simon Fraser University\\
Burnaby, Canada\\
oschulte@cs.sfu.ca}
}

% conference papers do not typically use \thanks and this command
% is locked out in conference mode. If really needed, such as for
% the acknowledgment of grants, issue a \IEEEoverridecommandlockouts
% after \documentclass

% for over three affiliations, or if they all won't fit within the width
% of the page, use this alternative format:
% 
%\author{\IEEEauthorblockN{Michael Shell\IEEEauthorrefmark{1},
%Homer Simpson\IEEEauthorrefmark{2},
%James Kirk\IEEEauthorrefmark{3}, 
%Montgomery Scott\IEEEauthorrefmark{3} and
%Eldon Tyrell\IEEEauthorrefmark{4}}
%\IEEEauthorblockA{\IEEEauthorrefmark{1}School of Electrical and Computer Engineering\\
%Georgia Institute of Technology,
%Atlanta, Georgia 30332--0250\\ Email: see http://www.michaelshell.org/contact.html}
%\IEEEauthorblockA{\IEEEauthorrefmark{2}Twentieth Century Fox, Springfield, USA\\
%Email: homer@thesimpsons.com}
%\IEEEauthorblockA{\IEEEauthorrefmark{3}Starfleet Academy, San Francisco, California 96678-2391\\
%Telephone: (800) 555--1212, Fax: (888) 555--1212}
%\IEEEauthorblockA{\IEEEauthorrefmark{4}Tyrell Inc., 123 Replicant Street, Los Angeles, California 90210--4321}}

% use for special paper notices
%\IEEEspecialpapernotice{(Invited Paper)}

% make the title area
\maketitle

\begin{abstract} This paper is based on a previous publication~\cite{Riahi2015}. Our work extends exception mining and outlier detection to the case of object-relational data. Object-relational data represent a complex heterogeneous network~\cite{Gao2010}, which comprises objects of different types, links among these objects, also of different types, and attributes of these links. 
%In relational structures, an individual object may display a complex pattern of probabilistic associations among its attributes, the attributes of linked objects, and the attributes of its links. These relationships represent a complex heterogeneous network of objects. 
This special structure prohibits a direct vectorial data representation. We follow the well-established Exceptional Model Mining framework, which leverages machine learning models for exception mining: A object is exceptional to the extent that a model learned for the object data differs from a model learned for the general population. Exceptional objects can be viewed as outliers. We apply state-of-the-art probabilistic modelling techniques for object-relational data that construct a graphical model (Bayesian network), which compactly represents probabilistic associations in the data. A new  metric, derived from the learned object-relational model, quantifies the extent to which the individual association pattern of a potential outlier deviates from that of the whole population. The metric is based on {\em the likelihood ratio} of two parameter vectors: One that represents the population associations, and another that represents the individual associations. 
%The likelihood ratio can be improved for outlier detection by applying two transformations: (1) a mutual information decomposition, and (2) replacing log-likelihood differences by log-likelihood distances. 
Our method is validated on synthetic datasets and on real-world data sets about soccer matches and movies. Compared to baseline methods, our novel transformed likelihood ratio achieved the best detection accuracy 
%(AUC) 
%true positive/negative rates 
on all datasets. 
\end{abstract}
% IEEEtran.cls defaults to using nonbold math in the Abstract.
% This preserves the distinction between vectors and scalars. However,
% if the conference you are submitting to favors bold math in the abstract,
% then you can use LaTeX's standard command \boldmath at the very start
% of the abstract to achieve this. Many IEEE journals/conferences frown on
% math in the abstract anyway.

% no keywords

% For peer review papers, you can put extra information on the cover
% page as needed:
% \ifCLASSOPTIONpeerreview
% \begin{center} \bfseries EDICS Category: 3-BBND \end{center}
% \fi
%
% For peerreview papers, this IEEEtran command inserts a page break and
% creates the second title. It will be ignored for other modes.
\IEEEpeerreviewmaketitle

\section{Introduction: Exception Mining for Relational Data} 
Exception mining is an important data analysis task in many domains. For relational data, exception mining supports outlier detection, where statistical deviations are viewed as due to a node or entity being genuinely exceptional, rather than due to statistical noise in the data. Statistical approaches to unsupervised exception/outlier detection are based on a generative model of the data~\cite{aggarwal2013}. 
The generative model represents normal behavior. An individual object is deemed an outlier if  the model assigns sufficiently low likelihood to generating it. 
Following the well-established Exceptional Model Mining framework \cite{Duivesteijn2016}, we propose a new method for extending statistical  outlier detection to the case of object-relational data using a novel likelihood-ratio comparison for generative probabilistic models. 

The object-relational data model is one of the main data models for structured data~\cite{Koller1997}. The main 
characteristics of objects that we utilize in this paper are the following. (1) {\em Object Identity.} Each object has a unique identifier that is the same across contexts. For example, a player has a name that identifies him in different matches. (2) {\em Class Membership.} An object is an instance of a class, which is a collection of similar objects. Objects in the same class share a set of attributes. For example, van Persie is a player object that belongs to the class striker, which is a subclass of the  class player. Note that this use of the term ``class'' is different from the machine learning sense of ``class'' as a prediction target. (3) {\em Object Relationships.} Objects are linked  to other objects. Both objects and their links have attributes. A common type of object relationship is a component relationship between a complex object and its parts.
For example, a match links two teams, and each team comprises a set of players for that match. A difference between relational and vectorial data is therefore that an individual object is characterized not only by a list of attributes, but also by its links and by attributes of the object linked to it. We refer to the substructure comprising this information as the {\em object data}. Equivalent terms are ``egonet'' from network analysis \cite{Akoglu2015} and ``interpretation'' \cite{Maervoet2012}. Relational outlier detection aims to identify objects whose data differ from the general population or class. Our approach to this problem leverages statistical-relational model discovery, as follows.

\paragraph{Approach} 
A class-model Bayesian network (BN) structure is learned with data for the entire population. The nodes in the BN represent attributes for links, of multiple types, and attributes of objects, also of multiple types. To learn the BN model, we apply techniques from statistical-relational learning, a  recent field that combines AI and machine learning \cite{SRL2007,Schulte2012,Domingos2009}. 
%The BN provides dimensionality reduction, in the sense that it leverages independencies to represent the data distribution with exponentially fewer parameters than a non-factorized parametrization. 
Given a set of parameter values and an input database, it is possible to compute a {\em class model likelihood} that quantifies how well the BN fits the object data. The class model likelihood uses BN parameter values {\em estimated from the entire class data.} This  is a relational extension of the standard log-likelihood method for i.i.d. vectorial data, which uses the likelihood of a data point as its outlier score. %This can be adapted for object-relational data as follows.
%The Bayes net structure represents the normal pattern of associations among links and attributes  by the well-known d-separation criterion: Two nodes are probabilistically independent if they are d-separated. 
 While the class model likelihood is a good baseline score, it can be improved by comparing it to {\em the object model likelihood}, which uses BN parameter values {\em estimated from the object data.}
The {\em model log-likelihood ratio} (LR) is the log-ratio of the object model likelihood to the class model likelihood. This ratio quantifies how the probabilistic associations that hold in the general population deviate from the associations in the object data substructure.
While the 
likelihood ratio discriminates relational outliers better than the class model likelihood alone, it can be improved further by applying two transformations: (1) a mutual information decomposition, and (2) replacing log-likelihood differences by log-likelihood distances. We refer to the resulting novel score as the {\em log-likelihood distance}.
\paragraph{Evaluation} \label{sec:eval} Our code and datasets are available on-line at \cite{url}.
Our performance evaluation follows the design of previous outlier detection studies~\cite{Gao2010,aggarwal2013},
%~\cite{Cansado2008, Muller2012},
%\footnote{review} 
where the methods are scored against a test set of known outliers.  
%and case studies assess their output on specific cases. 
%
We use three synthetic and two real-world datasets, from the UK Premier Soccer League and the Internet Movie Database (IMDb). On the synthetic data we have known ground truth. For the real-world data, we use a one-class design, where one object class is designated as normal and objects from outside the class are the outliers. For example, we compare goalies as outliers against the class of strikers as normal objects. 
%Given ground truth, an outlier detection method can be scored in terms of true positives and negatives, summarized using AUC~\cite{Muller2012}.\footnote{review} 
%Comparisons outlier scores include the class model likelihood, our novel log-likelihood distance, and likelihood-based scores intermediate to these two. 
%Previous outlier analysis for similar structured data  \cite{Breunig2000} used a preprocessing step where the structured data are converted to vectorial data that represent atomic objects. This conversion is usually done by aggregation, which tends to lose information.
%, for instance using counts as attributes in an attribute vector. After aggregation, we evaluate 
%After aggregation, standard outlier detection methods for independent data points can be applied; we use three as baseline methods for our evaluation ($\outrank$, $\lof$, and $\knn$).
On all datasets, the log-likelihood distance metric achieves the best detection accuracy compared to baseline methods. 

We also offer case studies where we assess whether individuals that our score ranks as highly unusual in their class are  indeed unusual. 
%This assessment looks at the object profile data of these individuals. 
The case studies illustrate that our outlier score is {\em easy to interpret}, because the Bayesian network provides a sum decomposition of the data distributions by features. Interpretability is very important for users of an outlier detection method as there is often no ground truth to 
evaluate outliers suggested by the method. %With regards to the cost of computing the divergence outlier score, we show that a Bayesian network representation of the object distributions can speed up the computation by orders of magnitude.

\begin{figure}[t]
\centering
\includegraphics[width=0.4\textwidth]{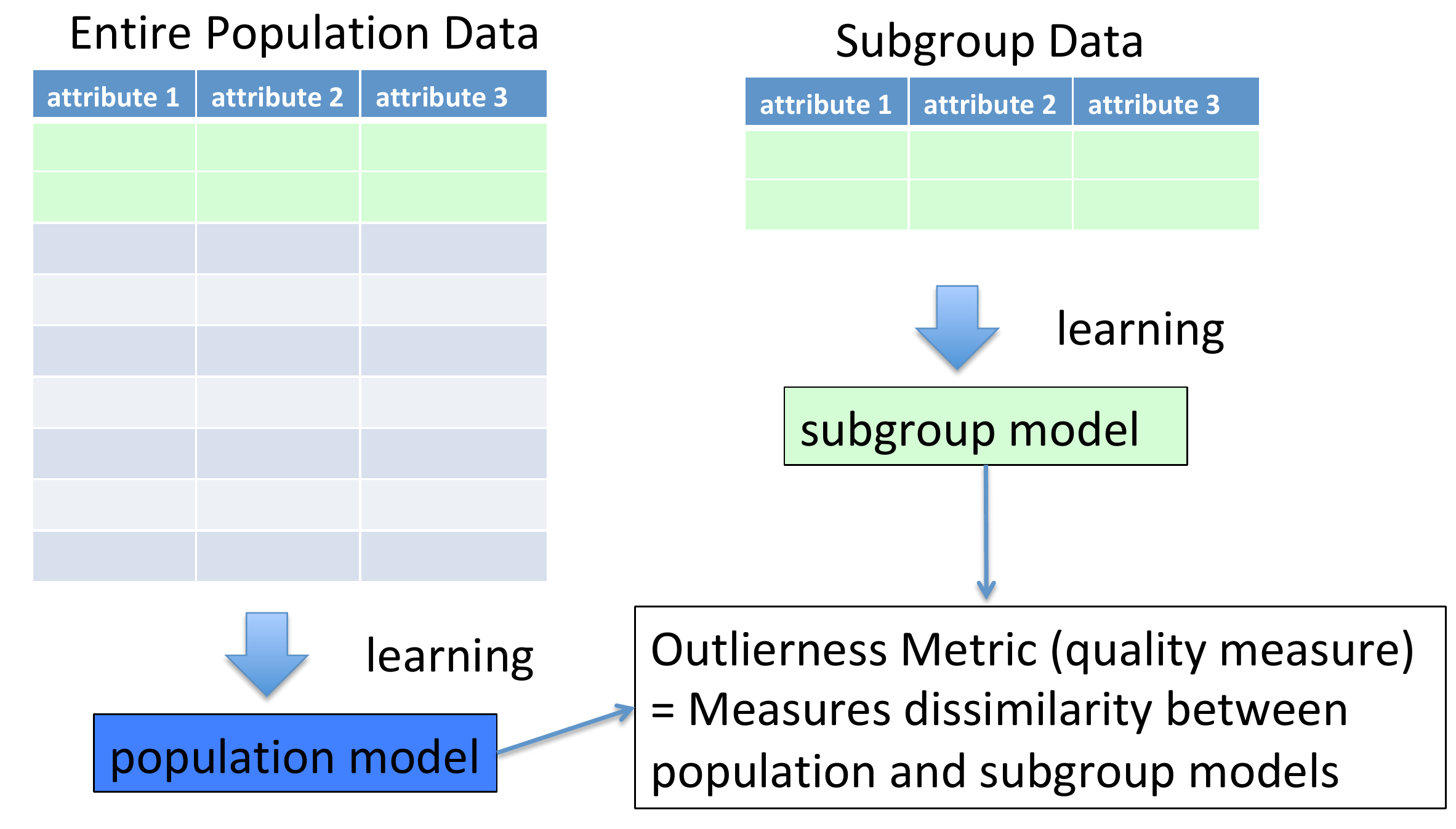}
\caption{A general schema for Exceptional Model Mining  for propositional data
\label{fig:emm}}
\end{figure}

\paragraph{Related Work} Section~\ref{sec:related} discusses the relationship to related work in detail. Our approach applies the exceptional model mining (EMM) framework~\cite{Duivesteijn2016} to multi-relational data. Figure~\ref{fig:emm} illustrates the EMM schema. 
The EMM framework leverages the extensive work on model learning in machine learning for exception mining: A subgroup is exceptional to the extent that a model learned from data for the subgroup deviates from a model learned for the general population. A computational method for measuring this extent is called a quality measure; we also refer to it as an outlierness metric. For a given model type, finding an appropriate quality measure for quantifying exceptionality is the main research question in EMM. The EMM framework allows us to leverage the extensive work on statistical-relational model learning for exception mining in multi-relational data. Compared to previous EMM models, the novelty of our work is as follows. 1) EMM has so far been developed only for propositional i.i.d. data, not relational data. Accordingly EMM has not been applied with SRL models. 2) In the propositional i.i.d. setting, each object is represented by a single data row, and it is meaningless to learn a model for a single object. Instead, EMM is applied to identify exceptional subgroups of objects. With relational data, each object is represented by its own dataset (egonet, interpretation), and it is meaningful to apply EMM to identify single exceptional objects. 
Compared to previous relational outlier detection work, our model-based approach is novel in that it neither summarizes the object data by a feature set (as in the Oddball system, see \cite{Akoglu2015}) nor looks for rules that exceptional objects violate (e.g. \cite{Maervoet2012}). 

\paragraph{Contributions} Our main contributions may be 
 summarized as follows.

\begin{enumerate} 
	\item The first EMM approach to outlier detection for structured data that is based on a probabilistic model. 
	\item A new outlier score based on a novel model likelihood comparison, the log-likelihood distance.  % \footnote{Commented paper organization}
	\end{enumerate}
% \vspace{-0.5cm} 

\paragraph{Paper Organization} We review background about Bayesian networks for relational data. Then we describe how we apply the EMM framework to multi-relational data. We introduce a novel log-likelihood distance outlier score as the quality or outlierness metric. After presenting the details of our approach, we review related work. Empirical evaluation compares model-based and aggregation-based approaches to relational outlier detection, with respect to three synthetic and three real-world problems.

%\paragraph{Paper Organization} We review background about our main model class, Bayesian networks for relational data, including the definition of the model likelihood function. Then we introduce and analyze our novel log-likelihood distance outlier score, and other baseline likelihood-based scores. Empirical evaluation compares likelihood-based and aggregation-based approaches to relational outlier detection, with respect to three synthetic and three real-world problems. We report detection accuracy and qualitative case studies. Outlier detection is a densely researched field so we end with an extensive discussion of related work.

\section{Background: Bayesian Networks for Relational Data}
We adopt  
 the Parametrized Bayes net (PBN) formalism \cite{Poole2003} that combines Bayes nets with logical syntax for expressing relational concepts. EMM is an inclusive framework and can in principle be applied with other SRL models, such as Markov Logic networks \cite{Domingos2009}. We worked with PBNs because i) they offer the most scalable structure learning methods \cite{Schulte2012a} to support our larger datasets, and ii) the PBN conditional probability parameters can be easily interpreted, which means that the resulting exceptionality metrics can be easily interpreted  (see Section~\ref{sec:eval} below).

\subsection{Bayesian Networks}

A {\bf Bayesian Network (BN)} is a directed acyclic graph (DAG) whose nodes comprise a set of random variables \cite{Pearl1988}. Depending on context, we interchangeably refer to the nodes  and variables of a BN. Fix a set of variables $\Features = \{\feature_{1},\ldots,\feature_{n}\}$. 
%These are attributes of objects, which can and typically do belong to different classes. In statistical terms, each attribute defines a random variable. 
The possible values of $\feature_{i}$ are enumerated as $\{\nodevalue_{i1},\ldots,\nodevalue_{i\states_{i}}\}$. The notation $P(\feature_{i} = \nodevalue)\equiv P(\nodevalue)$ denotes the probability of variable $\feature_{i}$ taking on value $\nodevalue$. We also use the vector notation $P(\Features = \set{\nodevalue}) \equiv P(\set{\nodevalue})$ to denote the joint probability that each variable $\feature_{i}$ takes on value $\set{\nodevalue}_{i}$.

The conditional probability parameters of a Bayesian network specify the distribution of a child node given an assignment of values to its parent node. For an assignment of values to its nodes, a BN defines the joint probability as the product of the conditional probability of the child node value given its parent values, for each child node in the network. This means that the log-joint probability can be {\em decomposed} as the node-wise sum

\begin{equation} \label{eq:bn}
\ln P(\Features = \set{\nodevalue};\model,\parameters) = \sum_{i=1}^{n} \ln \parameter(\set{\nodevalue}_{i}|\set{\nodevalue}_{\parents_{i}})
\end{equation}

where $\set{\nodevalue}_{i}$ resp. $\set{\nodevalue}_{\parents_{i}}$ is the assignment of values to node $\feature_{i}$ resp. the parents of $\feature_{i}$ determined by the assignment $\set{\nodevalue}$. 
%The function $\ln$ is the binary logarithm base 2. 
To avoid difficulties with $\ln(0)$, here and below we assume that joint distributions are positive everywhere. Since the parameter values for a Bayes net define a joint distribution over its nodes, they therefore entail a marginal, or unconditional, probability for a single node. We denote the \textbf{marginal probability} that node $\feature$ has value $\nodevalue$ as $P(\feature = \nodevalue;\model,\parameters) \equiv \parameter(\nodevalue)$.

\begin{figure}[t]
	\centering
	\includegraphics[width=0.4\textwidth] 
	{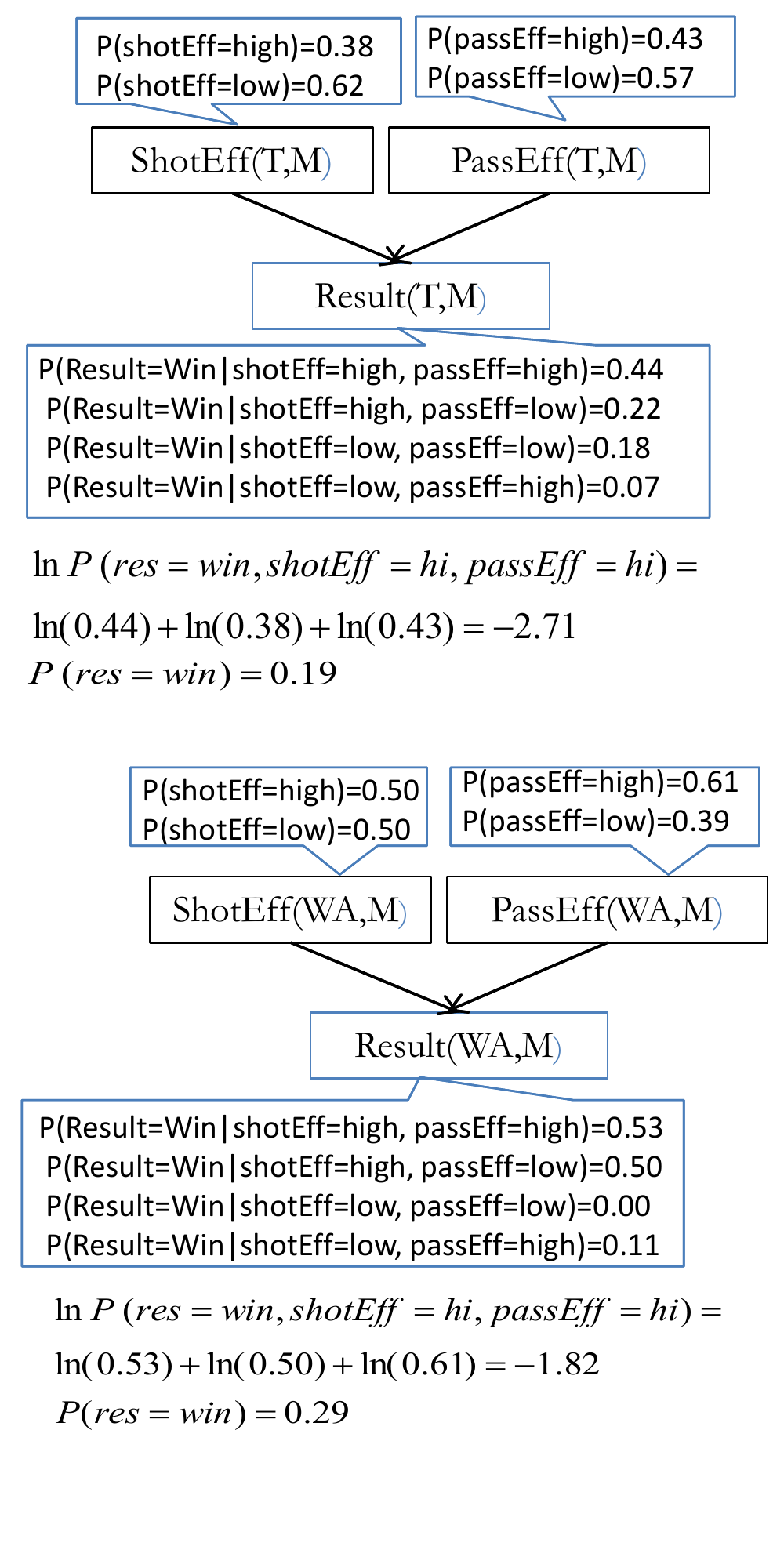}
	\caption{Example of joint and marginal probabilities computed from a toy Bayesian network structure. The parameters were estimated from the  Premier League dataset. (Top): A class model Bayesian network $\model_{\class}$ for all teams with class parameters $\parameters_{\class}$. (Bottom): The same Bayesian network structure with object parameters $\parameters_{\object}$ learned for Wigan Athletics ($T = WA$). Our model-based methods outlier scores compare the data likelihood of the class parameters and the object parameters.
		%We show only the Markov blanket of the Results node to simplify. 
		\label{fig:bns}
		}
\end{figure}

\paragraph{Example.} Figure~\ref{fig:bns} shows an example of a Bayesian network and associated joint and marginal probabilities.
%
%\footnote{\textbf{Sarah}: add marginal probability P(res=win) for object parameters}

\begin{figure}[t]
\centering
\includegraphics[width=0.4\textwidth]{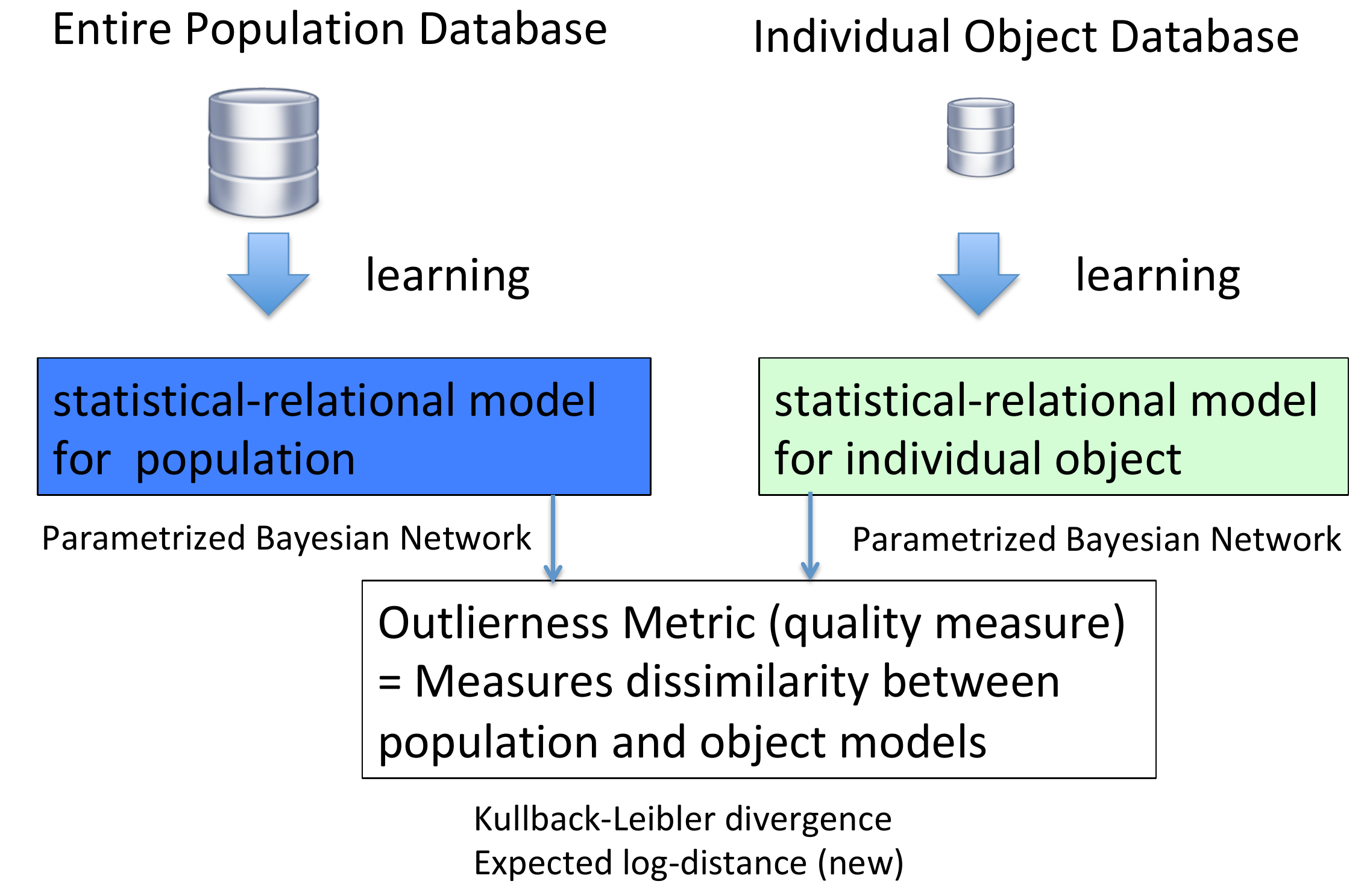}
\caption{The EMM approach for statistical-relational models. The model class we utilize in this paper are Parametrized Bayesian networks, with a log-linear likelihood function. As outlierness metrics we consider the standard Kullback-Leibler divergence, and a novel divergence introduced in this paper.
\label{fig:flow}}
\end{figure}
%				\begin{figure}
%				\centering
%				\resizebox{0.9\textwidth}{!}{
%				
%				\subfigure{
%				  \includegraphics[height=70mm, width=70mm] {figures/TPR-Synthetic.pdf}
%				}
%				\subfigure{
%				  \includegraphics[height=70mm, width=70mm] {figures/TPR-All.pdf}
%				
%				 }
%				 }
%				
%				\caption{Comparison of Object Outlier Metrics}
%				\label{fig:synthetic}
%				\end{figure}
%				
				
%				\begin{figure}
%					\centering
%					\begin{subfigure}{0.4\textwidth}
%						\includegraphics[width=1\linewidth]{figures/TPR-All.pdf}
%						\caption{}
%						\label{fig:Ng1} 
%					\end{subfigure}
%					
%					\begin{subfigure}{0.4\textwidth}
%						\includegraphics[width=1\linewidth]{figures/TPR-All.pdf}
%						\caption{}
%						\label{fig:Ng2}
%					\end{subfigure}
%					
%					\caption[Two numerical solutions]{(a) Numerical solutions for the small-time system 
%						with a constant-curvature body shape showing the scaled leading-order veritcal 
%						reaction force $N_0$ versus the scaled body mass $M$ for various values of $g$. 
%						Again, $I=M$ for definiteness and $A=0.7$. (b) As for (a) but over a wider range of 
%						values of $M,I$.}
%				\end{figure}
\subsection{Relational Data}

%We apply the learn-and-join algorithm (LAJ), which is the state-of-the-art Bayes net learning method for relational data. The LAJ algorithm takes as input a relational database and outputs a Bayes net using the functor notation due to Poole \cite{Poole2003}. We briefly review this notation.

%\paragraph{Functor Terms} 
 We adopt a functor-based notation for combining logical and statistical concepts~\cite{Poole2003,Kimmig2014}.
A functor is a function or predicate symbol. Each functor has a set of values (constants) called the \textbf{domain} of the functor. The domain of a \textbf{predicate} is $\{\true,\false\}$. Predicates are usually written with uppercase Roman letters, other terms with lowercase letters.
A predicate of arity at least two is a \textbf{relationship} functor. Relationship functors specify which objects are linked. Other functors represent \textbf{features} or \textbf{attributes} of an object or a tuple of objects (i.e., of a relationship).
A \textbf{population} is a set of objects. 
A \textbf{term} is of the form $f(\term_{1},\ldots,\term_{k})$ where $\functor$ is a functor %(either a function symbol or a predicate symbol) 
and each $\term_{i}$ is a first-order variable or a constant denoting an object. A term is \textbf{ground} if it contains no first-order variables; otherwise it is a first-order term. In the context of a statistical model, we refer to first-order terms as \textbf{Parametrized Random Variables} (PRVs) \cite{Kimmig2014}. 
%A term whose range are the truth values $\{\true,\false\}$ is a \textbf{predicate}. 
%Predicates are usually written with uppercase Roman letters, other term with lowercase letters.
%The grounding concept represents moving from the population-level  to the object level. 
A \textbf{grounding} replaces each first-order variable in a term by a constant; the result is a ground term. A grounding may be applied simultaneously to a set of terms.  A relational database $\D$ specifies the values of all ground terms, which can be listed in data tables. 
%In machine learning terminology, the data tables are contingency tables that represent sufficient statistics or event counts.

Consider a joint assignment 
$P(\Features = \set{\nodevalue})$ of values to a set of PRVs $\Features$. The {\em grounding space} of the PRVs is the set of all possible grounding substitutions, each applied to all PRVs in $\Features$. The {\em count} of groundings that satisfy the assignment with respect to a database $\D$ is denoted by $\grounds_{\D}(\Features = \set{\nodevalue})$. The \textbf{database frequency} $P_{\D}(\Features = \set{\nodevalue})$ is the grounding count divided by the number of all possible groundings.
%, that is, the size of the grounding space.
		
\emph{Example.} \label{sec:example}
The Opta dataset represents information about premier league data %\cite{opta-original} 
(Sec.~\ref{sec:real-world-data}). 
%Using the functor notation, the data
%format can be represented as follows. 
The basic populations are teams, players, matches, with 
corresponding first-order variables $\team, \player, \match$. Table~\ref{table:data} specifies values for some ground terms. The first three column headers show first-order variables ranging over different populations. The remaining columns represent features. Table~\ref{table:counts} illustrates grounding counts. Counts are based on the 2011-2012 Premier League Season. We count only groundings $(\it{team},\it{match})$ such that $\it{team}$ plays in $\it{match}$. Each team, including Wigan Athletics, appears in 38 matches. The total number of team-match pairs is $38 \times 20 = 760$.

%Examples of terms include the following. 

%\begin{itemize}
%\item $\it{Appears\_Player}(\P,\M)$ indicates whether a player appeared in a match.
%\item $\it{Appears\_Team}(\T,\M)$ indicates whether a team played in a match.
%\item $\it{Team}(\P)$ returns the team of a player.
%\item $\it{Result}(\T,\M)$ denotes the result of a team in a match (win or lose).
%\item $\it{ShotEff}(\T,\M)$ denotes the shot efficiency of a team in a match (number of successful shots on target, per total number of shots).
%\item $\it{TimePlayed}(\P,\M)$ denotes the total time that a player played in a match.
%\end{itemize}

%\begin{table}[htbp]
%\caption{Examples of terms in the soccer dataset.}
%\centering
%\resizebox{0.5\textwidth}{!}{
%\begin{tabular}{|c|p{5cm}|}
%\hline
%Term&Meaning\\ \hline
%$\it{Appears\_Player}(\P,\M)$ & indicates whether a player appeared in a match.\\ \hline
%$\it{Appears\_Team}(\T,\M)$&indicates whether a team played in a match.\\ \hline
%$\it{Team}(\P)$& returns the team of a player.\\ \hline
%$\it{Result}(\T,\M)$ &denotes the result of a team in a match (win or lose).\\ \hline
%$\it{ShotEff}(\T,\M)$ &denotes the shot efficiency of a team in a match (number of successful shots on target, per total number of shots).\\ \hline
%$\it{TimePlayed}(\P,\M)$& denotes the total time that a player played in a match.\\ \hline
%\end{tabular}}
%\label{table:terms}
%\end{table}

\begin{table}[htbp]
\caption{Sample Population Data Table (Soccer). \label{table:data}}
\centering
\resizebox{0.5\textwidth}{!}{
\begin{tabular}{|c|c|l|c|c|c|c|}
\hline
 \multicolumn{1}{|l|}{MatchId \match} &
 TeamId \team & PlayerId \player& \multicolumn{1}{l|}{First\_goal(\player,\match)} 
& \multicolumn{1}{l|}{TimePlayed(\player,\match)} & 
\multicolumn{1}{l|}{ShotEff(\team,\match)}&result(\team,\match) \\ \hline
 117 & WA & McCarthy & 0 & 90 & 0.53&\it{win}\\ \hline
 148 & WA & McCarthy & 0 & 85 & 0.57&\it{loss}\\ \hline
 15 & MC & Silva & 1 & 90 & 0.59&\it{win}\\ \hline
  \ldots& \ldots &\ldots&\ldots&\ldots&\ldots&\\
\end{tabular}}
\caption{Sample Object Data Table, for team $\team = \it{WA}$. \label{table:individual}}
\resizebox{0.5\textwidth}{!}{
\begin{tabular}{|c|c|l|c|c|c|c|}
\hline
 \multicolumn{1}{|l|}{MatchId \match} &
 TeamId $\team = \it{WA}$ & PlayerId \player& \multicolumn{1}{l|}{First\_goal(\player,\match)} 
& \multicolumn{1}{l|}{TimePlayed(\player,\match)} & 
\multicolumn{1}{l|}{ShotEff(\it{WA},\match)}&result(\it{WA},\match) \\ \hline
 117 & WA & McCarthy & 0 & 90 & 0.53&\it{win}\\ \hline
 148 & WA & McCarthy & 0 & 85 & 0.57&\it{loss}\\ \hline
   \ldots& WA &\ldots&\ldots&\ldots&\ldots&\\
\end{tabular}}
\end{table}

A novel aspect of our paper is that we learn model parameters for specific objects as well as for the entire population. 
%To implement this, for each target object, we form 
The appropriate \textbf{object data table} is formed from the population data table by restricting the relevant first-order variable to the target object. 
For example, the object database for target Team $\it{Wigan Athletic}$, 
forms a subtable of the data table of Table~\ref{table:data} that contains only rows where 
TeamID = $\it{WA}$; see Table~\ref{table:individual}. In database terminology, an object database is like a view centered on the object. The object database is an individual-centered representation \cite{Flach1999a}.

%\begin{table} 
%	\captionsetup{singlelinecheck=off}
%			\caption[.]{\label{table:counts}Example of Grounding Count and Frequency for the conjunction \begin{displaymath} \it{passEff(T,M)=hi}, shotEff(T,M)=high, Result(T,M)=1.\end{displaymath}}
%			\centering
%			\resizebox{0.5\textwidth}{!}{
%				\begin{tabular}{|c|c|c|}
%					\hline
%					Database&Count or $\#_{D}(\Features = \set{\nodevalue})$&Frequency or $P_{D}(\Features = \set{\nodevalue})$\\\hline
%					Population&76& $76/760=0.10$\\\hline
%					Wigan Athletics&7&$7/38=0.18$\\\hline
%		
%					%Synthetic&40&280\\ \hline
%				\end{tabular}}
%			\end{table}
\begin{table} 
	\caption{Example of Grounding Count and Frequency in Premier League Data, for the conjunction $\it{passEff(T,M)=hi}, shotEff(T,M)=hi, Result(T,M)=win$.\label{table:counts}}
	\centering
	\resizebox{0.5\textwidth}{!}{
		\begin{tabular}{|c|c|c|}
			\hline
			Database&Count or $\#_{D}(\Features = \set{\nodevalue})$&Frequency or $P_{D}(\Features = \set{\nodevalue})$\\\hline
			Population&76& $76/760=0.10$\\\hline
			Wigan Athletics&7&$7/38=0.18$\\\hline
			
			%Synthetic&40&280\\ \hline
		\end{tabular}}
	\end{table}

%\begin{table} 
%	\captionsetup{singlelinecheck=off}
%			\caption[.]{\label{table:counts}Example of Grounding Count and Frequency for the conjunction \begin{displaymath} \it{passEff(T,M)=hi}, shotEff(T,M)=high, Result(T,M)=1.\end{displaymath} Counts are based on the 2011-2012 Premier League Season. We count only groundings $(\it{team},\it{match})$ such that $\it{team}$ plays in $\it{match}$. Each team, including Wigan Athletics, appears in 38 matches. The total number of team-match pairs is $38 \times 20 = 760$.
%			\label{MetricComputation}}
%			\centering
%			\resizebox{0.5\textwidth}{!}{
%				\begin{tabular}{|c|c|c|}
%					\hline
%					Database&Count or $\#_{D}(\Features = \set{\nodevalue})$&Frequency or $P_{D}(\Features = \set{\nodevalue})$\\\hline
%					Population&76& $76/760=0.10$\\\hline
%					Wigan Athletics&7&$7/38=0.18$\\\hline
%		
%					%Synthetic&40&280\\ \hline
%				\end{tabular}}
%				
%			\end{table}
			
%[Example of counts, frequencies]
% For simplicity only, suppose that the only matches and players in the season are those shown in Table~\ref{table:data}. Then the attribute value $\it{First\_goals} = 0$ occurs with frequency $1/2$ in the object distribution $P_{\it{si}}$ for Silca. In the player class distribution $P_{\it{Player}}$, the attribute value $\it{First\_goals} = 0$ occurs with frequency $4/5$. So Silva is somewhat less likely to score no goal than a randomly selected player.
\subsection{Bayesian Networks for Relational Data}

A \textbf{Parametrized Bayesian Network Structure} (PBN) is a Bayesian network structure  whose nodes are PRVs. 
%For most of the paper we refer to PBNs simply as Bayesian networks, and to PRVs simply as random variables. 
%The learn-and-join algorithm is the state-of-the-art Bayes net learning method for relational data, based on model search in the lattice of relationship joins \cite{Schulte2012}. The LAJ algorithm takes as input a relational database and outputs a Parametrized Bayes net structure.
%We review the algorithm very briefly; for further details please see \cite{Schulte2012}. 
%The algorithm builds a Bayes net for an entire database by level-wise search through the {\em lattice of relationship chains.} This is the lattice of relationship sets that are connected by shared first-order variables.
%%; see Figure~\ref{fig:lattice}.  
%%We describe the fundamental ideas of the algorithm; for further details please see \cite{Schulte2012}. 
%The user chooses a single-table Bayes net learner. The learner is applied to base population data tables. Then the learner is applied to data tables for relationship chains of size $s,s+1,\ldots$, with the constraint that the models for larger join tables inherit the absence or presence of learned edges from smaller join tables. 
%
The relationships and features in an object database define a set of nodes for Bayes net learning; see Figure~\ref{fig:bns}.
% shows a sample PBN.

%We use the following notation.

%\begin{figure}[ht]
%\centering
%   \includegraphics[width=0.3\textwidth] {lattice}
% \caption{Lattice of Premier League dataset.
% %We show only the Markov blanket of the Results node to simplify. 
% \label{fig:lattice}}
%\end{figure}

%\begin{itemize}
%\item $\D_{\population}$ is the database for the entire population; cf. Table~\ref{table:data}.
%\item $\D_{\target}$ is the restriction of the input database to the target object; cf. Table~\ref{table:object}. 
%\item $\M_{\population}$ is a model (e.g., Bayesian network) learned with $\D_{\population}$ as the input database; cf. Figure~\ref{fig:bns}(a).
%\item $\M_{\target}$ is an object model (e.g., Bayesian network) learned with $\D_{\target}$ as the input database; cf. Figure~\ref{fig:bns}(b).
%\end{itemize}
%

\subsubsection{Model Likelihood for Parametrized Bayesian Networks}

A standard method for applying a generative model assumes that the generative model represents normal behavior since it was learned from the entire population. An object is deemed an outlier if the model assigns sufficiently low likelihood to generating its features \cite{Cansado2008}. This likelihood method is an important baseline for our investigation.
%, so we define the likelihood function formally in this section. 
%The other outlier scores we consider can be viewed as improved variants of the likelihood approach. 
Defining a likelihood for relational data is more complicated than for i.i.d. data, because an object is characterized not only by a feature vector, but by an object  database.
% that lists the object's links and the attributes of linked entities. 
%For the model likelihood function $\lnlikelihood(\model,\D)$, where $\model$ denotes a Bayesian network, 
We employ the relational pseudo log-likelihood \cite{Schulte2011}, which can be computed as follows for a given Bayesian network  and database.

%\begin{eqnarray}
%\lnlikelihood(\model,\D) =   \sum_{i=1}^{n}\sum_{j=1}^{\states_{i}} \sum_{\parents_{i}}\\ P_{\D}(\feature_{i} = \nodevalue_{ij},\parents_{i})\ln \parameter_{\model}(\feature_{i} = \nodevalue_{ij}|\parents_{i})  \nonumber
%\end{eqnarray}

\begin{equation} \label{eq:likelihood}
\loglikelihood(\D,\model,\parameters) =   \sum_{i=1}^{n}\sum_{j=1}^{\states_{i}} \sum_{\parents_{i}}\P_{\D}(\nodevalue_{ij},\parents_{i})\ln \parameter(\nodevalue_{ij}|\parents_{i})  
\end{equation}

Equation~\eqref{eq:likelihood} represents the standard BN log-likelihood function for the object  data~\cite{Campos2006}, except that parent-child instantiation counts are standardized to be proportions \cite{Schulte2011}. The equation can be read as follows.

\begin{enumerate}
\item For each parent-child configuration, 
use the conditional probability of the child given the parent.
\item Multiply the logarithm of the conditional probability by the database frequency of the parent-child configuration. 
%The instantiation proportion is the number of instantiating groundings, divided by the total number of possible instantiations.
\item Sum this product over all parent-child configurations and all nodes. 
\end{enumerate}

%\begin{eqnarray}
%\lnlikelihood&   \sum_{i=1}^{n}\sum_{j=1}^{\states_{i}} \sum_{\parents_{i}} P_{\D}(\feature_{i} = \nodevalue_{ij},\parents_{i})\ln \parameter_{\model}(\feature_{i} = \nodevalue_{ij}|\parents_{i})
%\end{eqnarray}

Schulte proves that the maximum of the pseudo-likelihood ~\eqref{eq:likelihood} is given by the empirical database frequencies \cite[Prop.3.1.]{Schulte2011}. In all our experiments we use these maximum likelihood parameter estimates.

{\em Example.} The family configuration \begin{displaymath} \it{passEff(T,M)=hi}, shotEff(T,M)=hi, Result(T,M)=win\end{displaymath} contributes one term to the pseudo log-likelihood for the BN of Figure~\ref{fig:bns}. For the population database, this term is $0.1 \times \ln(0.44) =-0.08 $. For the  Wigan Athletics database, the term is $0.18 \times \ln(0.44) =-0.14 $. 

\section{EMM for Relational Data} \label{sec:metrics}

This section describes our approach to applying the EMM framework to relational data, using the following notation.
\begin{itemize}
\item $\D_{\Class}$ is the database for the entire class of objects; cf. Table~\ref{table:data}. This database defines the \textbf{class distribution} $P_{\Class} \equiv P_{\D_{\class}}$.
\item $\D_{\object}$ is the restriction of the input database to the target object; cf. Table~\ref{table:individual}. This database defines the \textbf{object distribution} $P_{\object} \equiv P_{\D_{\object}}$.
\item $\model_{\Class}$ is a model (e.g., Bayesian network) learned with $\D_{\population}$ as the input database; cf. Figure~\ref{fig:bns}(a).
\item $\parameters_{\Class}$ resp. $\parameters_{\object}$ are parameters learned for $\model_{\Class}$ using $\D_{\class}$ resp. $\D_{\object}$ as the input database.
\end{itemize}

Figure~\ref{fig:flow} illustrates these concepts and the system flow for computing an outlierness score. First, we learn a Bayesian network $\model_{\Class}$ for the entire population using a previous learning algorithm (see Section~\ref{sec:methods} below). We then evaluate {\em how well the class model fits the target object data.} For vectorial data, the  standard model fit metric %approach 
is the log-likelihood of the target {\em datapoint}. For relational data, the counterpart is the relational log-likelihood \eqref{eq:likelihood} of the target {\em database}:

\begin{equation} \label{eq:loglikelihood-score}
\loglikelihood(\D_{\object},\model_{\Class},\parameters_{\Class}).
\end{equation}

While this
%the log-likelihood~\eqref{eq:loglikelihood-score} 
is a good baseline outlier score, it can be improved by considering scores based on the likelihood ratio, or {\bf log-likelihood difference}:
%. The log-likelihood difference is defined by

\begin{equation} \label{eq:log-diff}
\lr(\D_{\object},\model_{\Class},\parameters_{\object}) \equiv \loglikelihood(\D_{\object},\model_{\Class},\parameters_{\object}) - \loglikelihood(\D_{\object},\model_{\Class},\parameters_{\Class}).
\end{equation}

The log-likelihood difference compares  how well the class-level parameters fit the object data, vs. how well the object parameters fit the object data. In terms of the conditional probability parameters, it measures how much the log-conditional probabilities in the class distribution differ from those in the object distribution. Note that this definition applies only for relational data where an individual is characterized by a substructure rather than a ``flat'' feature vector. Assuming maximum likelihood parameter estimation, $\lr$ is equivalent to the Kullback-Leibler divergence between the class-level and object-level parameters~\cite{Campos2006}. 
While the $\lr$ score provides more outlier information than the model log-likelihood, it can be improved further by two transformations as follows. (1) Decompose the joint probability into a single-feature component and a mutual information component. (2) Replace log-likelihood differences by log-likelihood distances. The resulting score is the \textbf{log-likelihood distance} ($\mid$), which is the main novel score we propose in this paper. Formally it is defined as follows for each feature $i$. The total score is the sum of feature-wise scores. Section~\ref{sec:divergence-examples} 
 below provides example computations.
%In other words, we assume that model comparison scores are decomposable, which is the case for Bayesian networks. 
%\small
%\begin{equation} \label{eq:mid-bn}
%\begin{aligned}
%%\sum_{j=1}^{\states_{i} P_{\object}(\feature_{i} = \nodevalue_{ij}) \left|\ln \frac{P_{\object}(\feature_{i} = \nodevalue_{ij})}\right| + \\
%%\sum_{i=1}^{n} \sum_{j=1}^{\states_{i}}P_{\object}(\feature_{i}=\nodevalue_{ij}) \ln \frac{P_{\object}(\feature_{i} = \nodevalue_{ij})}{P_{\Class}(\feature_{i} = \nodevalue_{ij})} +\\
%\mid_{i} =\sum_{j=1}^{\states_{i}} P_{\object}(\feature_{i} = \nodevalue_{ij}) \left|\ln \frac{P_{\object}(\feature_{i} = \nodevalue_{ij})}{P_{\Class}(\feature_{i} = \nodevalue_{ij})}\right|+\\
%\sum_{j=1}^{\states_{i}} \sum_{\parents_{i}} 
%P_{\object}(\feature_{i} = \nodevalue_{ij},\parents_{i})\\
%\left|\ln \frac{P_{\object}(\feature_{i} = \nodevalue_{ij}|\parents_{i})}{P_{\object}(\feature_{i})=\nodevalue_{ij}} - \ln \frac{P_{\Class}(\feature_{i} = \nodevalue_{ij}|\parents_{i})}{P_{\Class}(\feature_{i})=\nodevalue_{ij}}\right|.
%\end{aligned}
%\end{equation}
%\normalsize

%\small
\begin{equation} \label{eq:log-dist}
\begin{array}{l}
 %\sum_{j=1}^{\states_{i} P_{\object}(\feature_{i} = \nodevalue_{ij}) \left|\ln \frac{P_{\object}(\feature_{i} = \nodevalue_{ij})}\right| + \\
		%\sum_{i=1}^{n} \sum_{j=1}^{\states_{i}}P_{\object}(\feature_{i}=\nodevalue_{ij}) \ln \frac{P_{\object}(\feature_{i} = \nodevalue_{ij})}{P_{\Class}(\feature_{i} = \nodevalue_{ij})} +\\
		\mid_{i} =\sum_{j=1}^{\states_{i}} P_{\object}(\nodevalue_{ij}) \left|\ln \frac{\parameter_{\object}( \nodevalue_{ij})}{\parameter_{\Class}( \nodevalue_{ij})}\right|+\\
		\sum_{j=1}^{\states_{i}} \sum_{\parents_{i}} 
		P_{\object}( \nodevalue_{ij},\parents_{i})
		\left|\ln \frac{\parameter_{\object}( \nodevalue_{ij}|\parents_{i})}{\parameter_{\object}(\nodevalue_{ij})} - \ln \frac{\parameter_{\Class}( \nodevalue_{ij}|\parents_{i})}{\parameter_{\Class}(\nodevalue_{ij})}\right|. \end{array}
		\end{equation}
		%\normalsize
%\paragraph{Motivation} 
%
%We note that for a fixed object distribution $P_{\object}$, the log-likelihood distance is a proper distance metric between the class-level and the object-level parameters. 
%The $\mid$ has two components. 
The first sum is the \textbf{single-feature} component, where each feature is considered independently of all others. It computes the expected log-distance with respect to  the singe feature value probabilities between the object and the class models. 
%For example, the single-feature sum for the feature ``Goal" of Van Persie is 5/8. 
%
The second $\mid$ sum is the \textbf{mutual information component}, based on the mutual information among all features. It computes the expected log-distance between the object and the class models with respect to the mutual information of feature value assignments.
%measures the expected distance in 
%%{\em multi-variate mutual information} 
%{\em association component} between the object and the class distributions. 
Intuitively, the first sum measures how the models differ if we treat each feature in isolation. The second sum measures how the models differ in terms of how strongly parent and child features are associated with each other. 
%A standard result in information theory states that a joint distribution can, without loss of information, be decomposed into a univariate distribution and the multi-variate mutual information  \cite{Witten2005}. Considering how the distributions differ in each of these two terms therefore involves no loss of information. 

%$\mid$ is not symmetric between the object and the class distribution because it weights sum terms by the joint probability in  the object distribution. The lack of symmetry is a desirable feature because to assess whether an object is an outlier, we want to weight most the events that occur more frequently with the object. 

\subsection{Motivation} 
%The motivation for using log-distances %rather than log-differences 
%is that some log-differences are positive, some negative, and cancel each other out when their sign differs. Since our goal is to assess the distinctness of an object, {\em we do not want differences to cancel out.} Fundamentally, averaging differences is appropriate when considering costs, payoffs or utilities, but not appropriate when assessing the distinctness of an object. 
%
The motivation for the mutual information decomposition is two-fold. 

\noindent
(1) {\em Interpretability}, which is very important for outlier detection. The single-feature components are easy to interpret since they involve no feature interactions. Each parent-child local factor is based on the average relevance of parent values for predicting the value of the child node, where relevance is measured by $\ln (\parameter(\nodevalue_{ij}|\parents_{i})/\parameter(\nodevalue_{ij}))$. This relevance term  is basically the same as the widely used lift measure \cite{Tuffery2011}, therefore an intuitively meaningful quantity. The $\mid$ score compares how relevant a given parent condition is in the object data with how relevant it is in the general class. 

\noindent
(2) {\em Avoiding cancellations.} Each term in the log-likelihood difference \eqref{eq:log-diff} decomposes into a relevance difference and a marginal difference: 

\begin{equation} \label{eq:decompose}
\ln \frac{\parameter_{\object}( \nodevalue_{ij}|\parents_{i})}{\parameter_{\Class}( \nodevalue_{ij}|\parents_{i})} = \ln \frac{\parameter_{\object}(\nodevalue_{ij}|\parents_{i})}{\parameter_{\object}(\nodevalue_{ij})} - \ln \frac{\parameter_{\Class}( \nodevalue_{ij}|\parents_{i})}{\parameter_{\Class}(\nodevalue_{ij})} + \ln \frac{\parameter_{\object}( \nodevalue_{ij})}{\parameter_{\Class}( \nodevalue_{ij})}.
\end{equation}

These differences can have different signs for different child-parent configurations and cancel each other out; see Table~\ref{table:Formula}. Taking distances as in Equation~\ref{eq:log-dist} avoids this undesirable cancellation. Since our goal is to assess the distinctness of an object, {\em we do not want differences to cancel out.} The general point is that averaging differences is appropriate when considering costs, or utilities, but not appropriate for assessing the distinctness of an object. For instance, the average of both vectors (0,0) and (1,-1) is 0, but their distance is not.

\subsection{Comparison Outlier Scores} Our lesion study compares our log-likelihood distance  $\mid$ score to baselines that are defined by omitting a component of $\mid$. In this section we define these scores.
 %and provide a theoretical comparison. 
  The scores increase in sophistication in the sense that they apply more transformations of the log-likelihood ratio. 
%Our empirical comparison below indicates that the 
More sophisticated scores provide more information about outliers.   
%defines the scores formally.  
Table~\ref{table:Formula} defines local feature scores; the total score is the sum of feature-wise scores. All metrics are defined such that {\em a higher score indicates a greater anomaly.} The metrics are as follows. 

\begin{LaTeXdescription}
\item[Feature Divergence $\fd$] is the first  component of the $\mid$ score. It considers each feature independently (no feature correlations).
\item[Log-Likelihood Score \loglikelihood] is the standard model-based outlier detection score using data likelihood.
\item[Log-Likelihood Difference \lr] is the log-likelihood difference~\eqref{eq:log-diff} between the class-level and object-level parameters. %(Equation). 
\item[Log-Likelihood Difference with absolute value $|\lr|$] replaces differences in $\lr$ by distances.
\item[Log-Likelihood Difference with decomposition $\lr^{+}$] applies a mutual information decomposition to $\lr$.
\end{LaTeXdescription}

	\begin{table}
		\caption{Baseline Outlier Scores for Bayesian Networks
			\label{table:Formula}}
		\resizebox{0.5\textwidth}{!}{
			\begin{tabular}{|l|l|} \hline
					Method & Formula\\
					\hline
				
							$\fd_{i}	$&	$\begin{array}{l}\sum_{i=1}^{n}\sum_{j=1}^{\states_{i}} P_{\object}( \nodevalue_{ij}) \left|\ln \frac{\parameter_{\object}( \nodevalue_{ij})}{\parameter_{\Class}( \nodevalue_{ij})}\right|\end{array}	$\\ \hline
						
						$-\loglikelihood_{i}$& $  \begin{array}{l} -\sum_{i=1}^{n}\sum_{j=1}^{\states_{i}} \sum_{\parents_{i}} P_{\object}( \nodevalue_{ij},\parents_{i})\ln \parameter_{\Class}( \nodevalue_{ij}|\parents_{i})\end{array}$ \\ \hline
					$\lr_{i}$&$\begin{array}{l}  \sum_{j=1}^{\states_{i}} \sum_{\parents_{i}} P_{\object}( \nodevalue_{ij},\parents_{i})\ln \frac{\theta_{\object}(  \nodevalue_{ij}|\parents_{i})}{\theta_{\Class}( \nodevalue_{ij}|\parents_{i})}.  \end{array}$\\	\hline
					$|\lr_{i}|$& $\begin{array}{l}  \sum_{j=1}^{\states_{i}} \sum_{\parents_{i}} P_{\object}( \nodevalue_{ij},\parents_{i})|\ln \frac{\theta_{\object}(\nodevalue_{ij}|\parents_{i})}{\theta_{\Class}( \nodevalue_{ij}|\parents_{i})}|. \end{array}$ \\	\hline
					$\lr_{i}^{+}$&$\begin{array}{l}  \sum_{j=1}^{\states_{i}} P_{\object}( \nodevalue_{ij}) \ln \frac{\parameter_{\object}( \nodevalue_{ij})}{\parameter_{\Class}( \nodevalue_{ij})}+
					\\ \sum_{j=1}^{\states_{i}} \sum_{\parents_{i}} 
					P_{\object}( \nodevalue_{ij},\parents_{i})
					\ln \frac{\theta_{\object}( \nodevalue_{ij}|\parents_{i})}{\parameter_{\object}(\nodevalue_{ij})} - \ln \frac{\theta_{\Class}( \nodevalue_{ij}|\parents_{i})}{\parameter_{\Class}(\nodevalue_{ij})}.  \end{array}$ \\ \hline
				
			\end{tabular} 
		}
	\end{table}

\begin{figure*}[htbp]
	\centering
	\resizebox{0.85\textwidth}{!}{
		\includegraphics%[width=0.3\textwidth] 
		{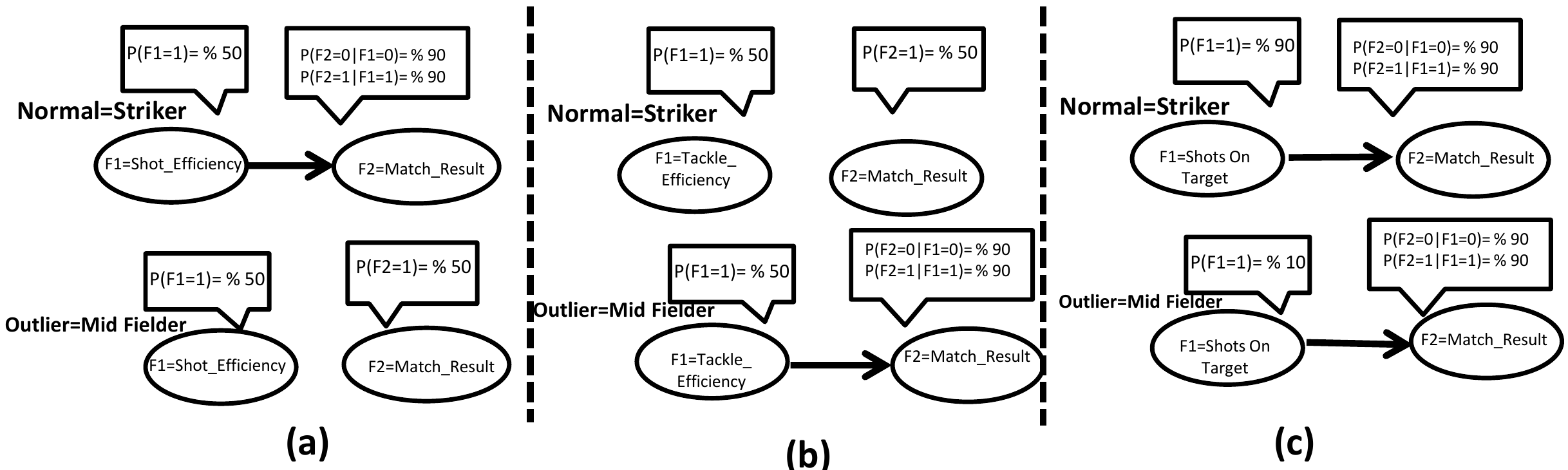}
	}
	\caption{Illustrative Bayesian networks. The networks are not learned from data, but hand-constructed to be plausible for the soccer domain. (a) High Correlation: Normal individuals exhibit a strong association between their features, outliers no association. Both normals and outliers have a close to uniform distribution over single features.
		% The outlier distribution misses a correlation that is present in the normal population. The single feature distributions are uniform in both distributions. 
		(b) Low Correlation: Normal individuals exhibit no association between their features, outliers have a strong association. Both normals and outliers have a close to uniform distribution over single features.
		% The outlier object exhibits a correlation that is not present in the normal population. The single attribute distributions are uniform in both distributions.
		(c) Single Attributes: Both normal and outlier individuals exhibit a strong association between their features. In normals, 90\% of the time, feature 1 has value 0. For outliers, feature 1 has value 0 only 10\% of the time. 
		%Correlations are the same, but the single feature distributions are not.
		\label{fig:synthetic-bns}}
\end{figure*}

\section{Examples} \label{sec:divergence-examples}

We provide three simple examples with only two features that illustrate the computation of the outlier scores. They are designed so that outliers and normal objects are easy to distinguish, and so that it is easy to trace the behavior of an outlier score.
%The examples also illustrate the concepts of attribute and correlation outlier. 
The examples therefore serve as thought experiments that bring out the strengths and weaknesses of model-based outlier scores. 
Figure~\ref{fig:synthetic-bns} describes the BN representation of the examples. Table~\ref{table:example} provides the computation of the scores. For intuition, we can think of a soccer setting, where each match assigns a value to each attribute $F_i, i =  1,2$ for each player. 
Scores for the $F_{2}$ feature are computed conditional on $F_{1} = 1$. Expectation terms are computed first for $F_{2} = 1$, then $F_{2} = 0$. 

The single feature distributions are uniform, so the feature component $\mid_{1}$ 
		is 0 for each node in both examples.
The table illustrates the undesirable cancelling effects in $\lr$. In the high correlation scenario~\ref{fig:synthetic-bns}(a), the outlier object has a lower probability than the normal class distribution of $\it{Match\_Result} = 0$ given that $\it{Shot\_Efficiency} = 1$. Specifically, 0.5 vs. 0.9. The outlier object exhibits a higher probability $\it{Match\_Result} = 1$ than the normal class distribution, conditional on $\it{Shot\_Efficiency} = 1$; specifically, 0.5 vs. 0.1. In line 1, column 2 of Table~\ref{table:example}  the log-ratios $\ln(0.5/0.9)$ and $\ln(0.5/0.1)$ therefore have different signs. In the low correlation scenario~\ref{fig:synthetic-bns}(b), the cancelling occurs in the same way, but with the normal and outlier probabilities reversed. 
The cancelling effect is even stronger for attributes with more than two possible values.

\begin{table}[hbpt]
	\caption{Example Computation of different outlier scores.
	% for the examples.
	% of Figure~\ref{fig:synthetic-bns}(a),(b). 
	 } 
		\label{table:example}
	
	%\captionsetup[table]{skip=10pt}
	
	\centering
	\resizebox{0.5\textwidth}{!}{
		\begin{tabular}{|c|p{3cm}|p{6cm}|l|}
			\hline
			Score&$F1=1$ Computation&$F2|F1=1$ Computation&Result\\ \hline
			$\lr$&\begin{tabular}{p{5cm}} $1/2 \ln(0.5/0.5)=0 $\end{tabular}&$1/4\ln(0.5/0.9)+ 1/4\ln(0.5/0.1)$&0.36\\ \hline
			%$ELD$&$1/2|\ln(0.5/0.5)|=0$&$1/4|\ln(0.5/0.9)|+1/4|\ln(0.5/0.1)|$&0.79\\ \hline
			$|\lr|$&$0$ (no parents)&\begin{tabular}{p{5cm}}$1/4 |\ln(0.5/0.5)-\ln(0.9/0.5)|+1/4 |\ln(0.5/0.5)-\ln(0.1/0.5)|$\end{tabular}&0.79\\ \hline
			$\fd$&$|\ln(0.5/0.5)|=0$&\begin{tabular}{p{5.5cm}}$1/2|\ln(0.5/0.5)| + 1/2|\ln(0.5/0.5)|$\end{tabular}&0\\ \hline
			$$\mid$$&$0+0$&$0.79+\fd$ 
			% O.S.0$
			&0.79
			%$1/4(|\ln(0.9/0.5)-\ln(0.5/0.5)|+|\ln(0.1/0.5)-\ln(0.5/0.5)|+2|\ln(0.5/0.5)|)=0.54$
			\\ \hline
		\end{tabular}}
		%\centering
		\resizebox{0.5\textwidth}{!}{
			\begin{tabular}{p{8cm}}
				Table V(a): High Correlation Case, Figure~\ref{fig:synthetic-bns}(a).
				%: The scores for the object and class BNs of Figure~\ref{fig:synthetic-bns}(a).
				
			\end{tabular}}

			\centering
			\resizebox{0.5\textwidth}{!}{
				\begin{tabular}{|l|p{3cm}|p{6cm}|l|}
					\hline
					Score&$F1=1$ Computation&$F2|F1=1$ Computation&Result\\ \hline
					$\lr$&$1/2\ln(0.5/0.5)=0$&$0.5 \cdot 0.9 \ln(0.9/0.5)+ 0.5 \cdot 0.1 \ln(0.1/0.5)$&0.26\\ \hline
					%$ELD$&$1/2|\ln(0.5/0.5)=0|$&$0.5 \cdot 0.9|\ln(0.9/0.5)|+0.5 \cdot 0.1|\ln(0.1/0.5)|$&0.50\\ \hline
					$|\lr|$&$0$ (no parents) &$0.5 \cdot 0.9 |\ln(0.9/0.5)-\ln(0.5/0.5)|+ 0.5 \cdot 0.1 |\ln(0.1/0.5)-\ln(0.5/0.5)| $&0.50\\ \hline
					$\fd$&$|\ln(0.5/0.5)|=0$&$1/2|\ln(0.5/0.5)| + 1/2|\ln(0.5/0.5)|$&0\\ \hline
					$\mid$&$0+0$&$0.5+\fd$
					% O.S. $0$
					&0.5\\ \hline
				\end{tabular}}
				\resizebox{0.5\textwidth}{!}{
					\begin{tabular}{p{8cm}}
						Table V(b): Low Correlation Case. Figure~\ref{fig:synthetic-bns}(b).
						 %The scores for the object and class BNs of Figure~\ref{fig:synthetic-bns}(b). 
						 
					\end{tabular}}
				\end{table}

\section{Related Work} \label{sec:related}
\begin{figure}
	\centering
	\includegraphics[width=0.5\textwidth] {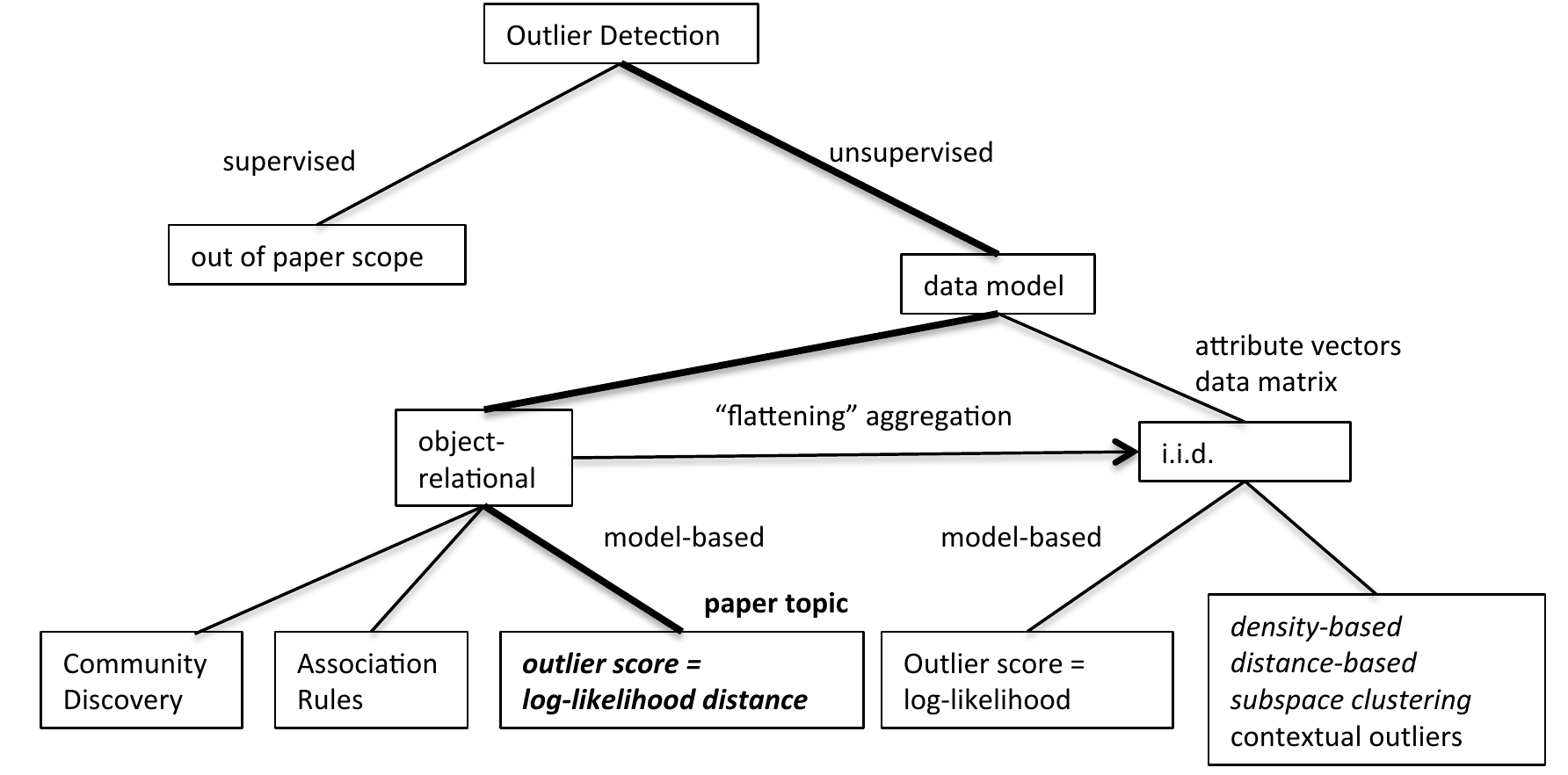}
	\caption{A tree structure for related work on outlier detection for structured data. A path specifies an outlier detection problem, the leaves list major approaches to the problem. Approaches in italics appear in experiments.
		\label{fig:novelty}}
\end{figure}

Outlier detection is a densely researched field, for a survey please see~\cite{aggarwal2013}.
Figure~\ref{fig:novelty} provides a tree picture of where our method is situated with respect to other outlier detection methods and other data models. 
%For an outlier detection survey please see~\cite{aggarwal2013}.
%\cite{Hodge2004,aggarwal2013}. 
Our method falls in the category of {\em unsupervised} statistical model-based approaches. To our knowledge, ours is the first model-based method tailored for object-relational data. Like other model-based approaches, it detects {\em global outliers.} Aggarwal \cite{aggarwal2013} defines a global outlier to be a data point that notably deviates from the rest of the population. We review relevant approaches from different data models, the most common atomic object model---where data is represented by vectors---and structured data models. Akoglu et al. provide an excellent recent survey of outlier detection in relational models~\cite{Akoglu2015}.\\

% using the XML, SQL, and OLAP formats.
%\vspace{-5mm}
\textit{a) Attribute Vector Data Model:}
%\footnote{\textbf{Sarah}: can you fix the silly d) for paragraph numbering?}By far most work on outlier detection considers atomic objects with flat feature vectors.
 By far most work on outlier detection considers atomic objects with flat feature vectors.
%, or nonhierarchical structures like time series. 
This leads to an impedance mismatch: 
The required input format for these outlier detection methods is a single data matrix, not a structured dataset. For example, one cannot provide a relational database as input. This mismatch is not simply a question of choosing a file format, but instead reflects a different underlying data model: complex objects with both attributes and component objects vs. atomic objects with attributes only. 
It is possible to ``flatten'' structured data by converting it to unstructured feature vectors, for instance by using aggregate functions. 
%Flattening incurs some loss of information but allows us to apply the many feature vector methods.
%\cite{Elke2013}. 
We evaluated the aggregation approach in this paper by applying three standard methods for outlier detection.
%for three major approaches to outlier detection: distance-based, density-based, and subspace clustering. 
%
%Subspace clustering methods (e.g., \cite{Muller2012,Kriegel2009}) are similar to our work in the sense that they aim to decompose a complex data space. They find complex deviations that are noticeable only in a data subspace. A common approach is to discover datapoints that show unexpected deviations in similar subspaces. Our approach instead develops a joint measure of how dissimilar the target object profile distribution is to the class distribution over the entire data space. Given that object and class distributions are represented by an object-oriented Bayesian network \cite{Koller1997}, the network structure defines subspaces. The joint divergence measure {\em mathematically decomposes} into subspace measures that quantify how dissimilar the target object profile distribution is in the subspaces defined by the network, compared to the class distribution in the same subspace.

Work on atomic contextual  outliers \cite{Tang2013} is like ours in that it considers the distinctness of a target individual from a reference class. A reference class is not specified for each object,
%as a property of the object, 
but is constructed as part of outlier detection. 
Our work could be combined with a class discovery approach by providing a score of how informative the inferred classes are. 
%\vspace{-5mm}

\textit{b) Structured Data Models:} We discuss related techniques in three types of structured data models: SQL (relational), XML (hierarchical), and OLAP (multi-dimensional). 

For relational data, many outlier detection approaches aim to discover rules that represent the presence of anomalous associations for an individual or the absence of normal associations \cite{Maervoet2012,Gao2010}. The survey by \cite{Novak2009} unifies within a general rule search framework related tasks such as exception mining, which looks for associations that characterize unusual cases, subgroup mining, which looks for associations  characterizing important subgroups, and contrast space mining, which looks for differences between classes. Another rule-based approach uses Inductive Logic Programming techniques \cite{Angiulli2007}.
%,Angiulli2009}.
While local rules are informative, they are not based on a global statistical model and do not provide a single outlier score for each individual. 

A latent variable approach in information networks ranks potential outliers in reference to the latent communities inferred by network analysis \cite{Gao2010}. Our model aggregates information from entities and links of different types, but does not assume that different communities have been identified.

Koh {\em et al.}~\cite{Koh2008} propose a method for hierarchical structures represented in XML document trees. Their aim is to identify feature outliers, not class outliers as in our work. Also, they use aggregate functions to convert the object hierarchy into feature vectors. Their outlier score is based on local correlations, and they do not construct a model.

The multi-dimensional data model defines numeric measures for a set of dimensions. 
%A seminal approach to exploring a multi-dimensional datacube was presented by Sarawagi {\em et al.}~\cite{Sarawagi1998}. 
%The object and the multi-dimensional data models are similar in the respect that both objects and dimensions are ordered in a hierarchy. However, 
The differences in the two data models mean that multi-dimensional outlier detection models~\cite{Sarawagi1998} do not carry over to object-relational outlier detection. (1) The object data model allows but does not require any numeric measures. In our datasets, all features are discrete. Nor do we assume that it is possible to aggregate numeric measures to summarize lower-level data at higher levels.  
(2) In scoring a potential outlier object, our method considers other objects {\em both} below and above the target object in the component hierarchy. OLAP exploration methods consider only cells below or at the same level as the target cell. For example, in scoring a player, our method would consider features of the player's team.  
Also, the $\mid$ outlier score of an object is not determined by the outlier scores of its components, in contrast to the approach of Sarawagi {\em et al.}.
% They use values such as the most unusual cell that is below a target cell.
(3) Our approach models a joint distribution over features, exploiting correlations among features. Most of the OLAP-based methods consider only a single numeric measure at a time, not a joint model.  

Statistical data cleaning methods are related to outlier detection, in that erroneous data may be detected as outliers (e.g., \cite{de2016bayeswipe}). Nonetheless, these data cleaning methods differ from our work in several ways. 1) Although they often originate in the database community, they are usually developed only for single-table propositional data, not relational data. (An exception is the ERACER system \cite{Mayfield2010}.) 2) Our work assumes that the data is (mainly) correct, and identifies exceptional identities for the given data. 3) Data cleaning methods focus on unusual values or tuples (e.g. a mistaken rating for a movie by a user), not exceptional subdatabases or egonets.

				\section{Experimental Design}
%Details about our systems and algorithms that we use from previous work may be found in Section~\ref{sec:details}.  
All the experiments were performed on a 64-bit Centos machine with 4GB RAM and an Intel Core i5-480 M processor. The likelihood-based outlier scores were computed with SQL queries using JDBC, JRE 1.7.0. and MySQL Server version 5.5.34.
We describe the datasets used in our experiments.
				
				\subsection{Synthetic Datasets}
				
				We generated three synthetic datasets with normal and outlier players using the distributions represented in the three Bayesian networks of Figure~\ref{fig:synthetic-bns}. 
				%The features $F_{1}$ and $F_{2}$ represent two features of players. 
				Each player participates in 38 matches, similar to the real-world data. The main goal of designing synthetic experiments is to test the methods on  easy to detect outliers. Each match assigns a value to each feature $F_i, i =  1,2$ for each player. 
				\begin{LaTeXdescription}
				\item\textbf{High Correlation} See Figure~\ref{fig:synthetic-bns}(a).
				\item\textbf{Low Correlation} See Figure~\ref{fig:synthetic-bns}(b).
				\item\textbf{Single features} See Figure~\ref{fig:synthetic-bns}(c).
				\end{LaTeXdescription}

				%These datasets are designed so that outliers and normal objects are easy to distinguish, and so that it is easy to trace the behavior of an outlier score.
				
				%\begin{description}
				%\item[High Correlation]  See Figure~\ref{fig:synthetic-bns}(a).
				%\item[Low Correlation]  See Figure~\ref{fig:synthetic-bns}(b).
				%\item[Single attributes] See Figure~\ref{fig:synthetic-bns}(c).
				%\end{description}
				%
				We used the $\it{mlbench}$ package in $\it{R}$ to generate synthetic features in matches, following these distributions for 240 normal players and 40 outliers. We followed the real-world Opta data in terms of number of normal and outlier individuals. The scores are used to rank all 280 players. 
				\subsection{Real-World Datasets} \label{sec:real-world-data}
				Data tables are prepared from Opta data~\cite{opta-original} and IMDb~\cite{IMDb-original}. Our datasets and code are available on-line~\cite{bib:jbnsite}.
				\paragraph{Soccer Data} 
The Opta data were released by Manchester City. 
It lists box scores, that is, counts of all the ball actions within each game by each player, for the 2011-2012 season. 
%The data consists of information about the actions of a single player in a given match 
%from 2011 to 2012. 
%Number of goals, passes, fouls, tackles, saves and blocks and also position 
%assigned to a player in a match are examples of the information associated with each player. [list]
%Information about the teams in a season, such as number of home wins, draws or away wins can be extracted by massaging the data. 
%[The information can be visualized as a heterogeneous network that links players to teams, and teams to matches. ]
For each player in a match, our data set contains eleven player features.
% like $\it{TimePlayed}(\P,\M)$.
For each team in a match, there are five features computed as player feature aggregates, as well as the team formation and the result (win, tie, loss). 
There are two relationships, $\it{Appears\_Player}(\P,\M)$, $\it{Appears\_Team}(\T,\M)$. 
%We store the data in a relational database, with a table for each base population and a table for each relationship.

				\paragraph{IMDB Data} 
The Internet Movie Database (IMDB) is an on-line database of information related to films, television programs, and video games.
The IMDB website offers a dataset containing information on cast, crew, titles, technical details and biographies into a set of compressed text files. 
We preprocessed the data like \cite{Peralta2007} to obtain a database with seven tables: one for each population and one for the three relationships $\it{Rated}(\user,\movie)$, $\it{Directs}(\director,\movie)$, and $\it{ActsIn}(\actor,\movie)$.
				
%				Table~\ref{table:Features} lists relationships and the number of features. 
%				\begin{table}[htbp]
%							\caption{Relationships and Example Features in Real-World Datasets.
%								%For relationships please see text.
%								\label{table:Features}}
%					\centering
%					\resizebox{0.3\textwidth}{!}{
%						\begin{tabular}{|l|c|l|}
%							\hline
%							Path/Object Type & \#Attributes&Example\\ \hline
%							\begin{tabular}{c}Player-Team-Match \end{tabular} & 11&ShotEff \\ \hline
%							\begin{tabular}{c}Team-Match \end{tabular} & 7&TeamFormation \\ \hline\hline
%							Actor & 2&Quality \\ \hline
%							Director & 2&AvgRevenue\\ \hline
%							Movie&5&Genre\\ \hline
%							User& 2&Occupation\\ \hline
%							User-Movie & 1&Rating \\\hline
%							Actor-Movie & 1&Cast\_Position\\\hline
%						\end{tabular}}
%				
%					\end{table}
%\footnote{Commented table of relationships and features}
%					\begin{LaTeXdescription}
%						\item[Soccer Data]
%						The Opta data were released by Manchester City. 
%						It lists all the ball actions within each game by each player, for the 2011-2012 season.
%						\item[IMDb Data]
%						The Internet Movie Database (IMDb) is an online database of information related to films, television programs, and video games.
%						The IMDb website offers a dataset containing information on cast, crew, titles, technical details and biographies in a set of compressed text files \cite{Peralta2007}.
%					\end{LaTeXdescription}
					
					%\subsection{Outlier and Contrast Classes.}
					
					In real-world data, there is no ground truth about which objects are outliers. To address this issue, we employ a one-class design: we learn a model for the class distribution, with data from that class only. Then we rank all individuals from the normal class together with all objects from a contrast class treated as outliers, to test whether an outlier score recognizes objects from the contrast class as outliers.
			%		distinguishes objects from the normal class from objects in the contrast class. 
					Table~\ref{MetricComputation} shows the normal and contrast classes for three different datasets.  In-class outliers are possible, e.g. unusual strikers are still members of the striker class. Our case studies describe a few in-class outliers. In the soccer data, we considered only individuals who played more than 5 matches out of a maximum 38. 
					%The maximum number of matches played is 38.
%
%					In this design, we are only looking for the objects that are clearly deviating from the majority of the data. We are aware of the 
					
										%
					%\begin{description}
					%\item[Strikers vs. Goalies] The generic model is learned with match data from all 150 strikers. The outlier test cases are match data for all 22 goalies.
					%\item[Midfielder vs. Striker]  The generic model is learned with match data from all 172 midfielders. The outlier test cases are match data for 70 randomly selected strikers.
					%\item[Drama vs. Comedy] The generic model is learned with data for all 400 drama movies and 80 randomly selected comedy movies.
					%\end{description}
					%Figure~\ref{fig:synthetic} shows the true outlier rates for the different outlier metrics.  {\em On all datasets, the Mutual Information Score $\mid$ separates the normal class from the contrast class better than the other methods.} The same is true for the true negative rate; see supplementary file.
					%This rate reflects only the number of contrast objects that are ranked highly. Therefore we cannot expect a true outlier rate of 100\% because the normal class may also contain genuine outliers. If a metric correctly ranks these within-class outliers more highly than contrast class members, its TOR decreases. 
					% that are not in the contrast some strikers, may be genuine outliers within the striker class.
					
					\begin{table}
							\caption{Outlier/normal Objects in Real-World Datasets. }
							\label{MetricComputation}
						\centering
						\resizebox{0.4\textwidth}{!}{
							\begin{tabular}{|c|c|c|c|}
								\hline
								Normal&\#Normal&Outlier&\#Outlier\\\hline
								Striker & 153 & Goalie&22\\ \hline
								Midfielder & 155 & Striker&74\\ \hline
								Drama & 197 & Comedy&47\\ \hline
								%Synthetic&40&280\\ \hline
							\end{tabular}}
						
						\end{table}
						
%						\subsection{Performance Score}
%					 Our experiments provide empirical evidence that $\mid$ works better than other scores for object outlier detection.
%						Our performance score for outlier rankings is the area under curve ($\auc$) of the well-established receiver operating characteristic $\roc$ curve.
%						~\cite{Fawcett2006}. 
%						This has been widely used to measure the performance of outlier ranking methods~\cite{Cansado2008, Muller2012}. The relationship between false positive rate (1- Specificity) and true positive rate (Sensitivity) is captured by the $\roc$ curve. Ideally, the best performance is achieved when we have the highest sensitivity and the highest specificity. 
%					
%						The maximum values for $\auc$ is 1.0 indicating a perfect ranking with 100\% sensitivity and 100\% specificity. In order to compute the $\auc$ value, we used the \textit{R} package \textit{ROCR}~\cite{RROCR2012}. Given a set of outlier scores, one for each object, this package returns an $\auc$ value. 
%					
																
						\subsection{Methods Compared}
						\label{sec:methods}
						We compare two types of approaches, and within each approach several outlier detection methods. The first approach evaluates the likelihood-based outlier scores described in Section~\ref{sec:metrics}. For relational Bayesian network structure learning we utilize the previous learn-and-join algorithm (LAJ), which is
				a state-of-the-art BN structure learning method for relational data \cite{Schulte2012}. The LAJ algorithm employs an iterative deepening strategy, which can be described as as search through a lattice of table joins. For each table join, different BNs are learned and the learned edges are propagated from smaller to larger table joins. 	For a full description, complexity analysis, and learning time measurements, please see \cite{Schulte2012}. 	We used the implementation of the LAJ algorithm due to its creators \cite{bib:jbnsite}. 
					
%						probabilistic scores applied directly to the  data structured as an object tree. 
						
						The second approach first ``flattens" the structured data into a matrix of feature vectors, then applies standard matrix-based outlier detection methods. We refer to such methods as \textbf{aggregation-based}
						%Flattening the object data allows applying the rich set of matrix-based outlier detection methods ``as is" 
						(cf. Figures~\ref{fig:novelty}). For example, this was the approach taken by Breunig {\em et al.} for identifying anomalous players in sports data \cite{Breunig2000}. Following their paper, for each continuous feature in the object data, we use the average over its values, and for each discrete feature, we use the occurrence count of each feature value in the object data. Aggregation 
						tends to lose information about correlations.
						%\cite{DavidJensen2002}. 
Our experiments address the empirical question of whether this loss of information affects outlier detection. 
						%The advantage to the aggregation approach is that after aggregating to preprocess the data, any matrix-based outlier detection method can be applied; see Figure~\ref{fig:flow}. 
						We evaluated three standard matrix-based outlier detection methods: Density-based $\lof$~\cite{Breunig2000}, distance-based $\knn$~\cite{Ramaswamy2000} and subspace analysis $\outrank$~\cite{Muller2012}.
						%\footnote{commented descriptions of these methods }. 
						These represent common, fundamental  approaches for vectorial data. 
					%	Subspace analysis is especially relevant to our study because 
						Like $\mid$, subspace analysis is sensitive to correlations among features. 
% Our experiments applied $\outrank$ with two subspace clustering models, \textit{PRO-CLUS} \cite{Muller2012} and \textit{DISH} \cite{Kriegel2007}.  
						We used the available implementation of all three data matrix methods from the state of the art data mining software \textit{ELKI} \cite{Elke2013}. We used \textit{PRO-CLUS} as the clustering function for $\outrank$, recommended by~\cite{Muller2012}.

						\section{Empirical Results}
						
						We present results regarding computational feasibility, 
						%$\auc$ 
						predictive performance, and case studies.
						
						\subsubsection{Computational Cost of the $\mid$ Score.}
%						
%						Table~\ref{table:Number of Parameters} shows that the Bayesian network representation provides a highly compact summary of the target class distribution: the number of probabilities that need to be evaluated for computing the probabilistic scores decreases by a factor of 
%						%\textbf{Sarah:fix dash}
%						$10^{4}$\--$10^{5}$ depending on the dataset. 
%						
						Table~\ref{table:LearningTime} shows that the computation of the $\mid$ value for a given target object is feasible. On average, it takes a quarter of a minute for each soccer player, and one minute for each movie. This includes the time for parameter learning from the object database.
Learning the class model BN takes longer, but needs to be done only once for the entire object class. 
						{\em The BN model 
%						is crucial for computational feasibility because it 
						provides a crucial %compact 
 low-dimensional representation of the 
						%joint 
						distribution information in the data.} Table~\ref{table:Number of Parameters} compares the number of terms required to compute the $\mid$ score in the BN representation to the number of terms in an unfactored representation with one parameter for each joint probability.
						
						\begin{table}[htbp]
						
							\begin{subtable}
								%\caption{Time for computing the $\mid$ metric. This includes the time for Bayes Net structure learning.\label{table:LearningTime}}
								\centering
									\caption{Time (min) for computing the $\mid$ score. 
								%This includes the time for Bayes Net structure learning.
				\label{table:LearningTime}}
								\resizebox{0.4\textwidth}{!}{
									\begin{tabular}{|c|l|l|} \hline
										Dataset& Class Model
										%    \begin{tabular}{c} Class \\Terms (min)\end{tabular}   & 
										%    \begin{tabular}{c} Object \\Terms (min)\end{tabular}   
										& Average per Object  \\ \hline
										Strikers vs. Goalies&4.14&0.25\\ \hline
										Midfielder vs. Goalies &4.02&0.25 \\ \hline
										Drama vs. Comedy &8.30&1.00\\ \hline
									\end{tabular} 
								}
							
							\end{subtable}

							\begin{subtable}
								%\caption{The Bayesian network representation helps to decrease the number of terms that represent the class and object distributions. 
								%For relationships please see text.
								%\label{table:Number of Parameters}}
								\centering
									\caption{The Bayesian network representation decreases the number of terms required for computing the $\mid$ score.
										%that represent the class and object distributions.
										\label{table:Number of Parameters}}
								\resizebox{0.4\textwidth}{!}{
									\begin{tabular}{|c|l|l|}
										\hline
										Dataset & \begin{tabular}{c} \#Terms \\Using BN\end{tabular}  & \begin{tabular}{c} \#Terms \\ without Using BN
										\end{tabular}\\ \hline
										Strikers vs. Goalies & 1,430&114,633,792\\ \hline
										Midfielders vs. Goalies & 1,376&43,670,016\\ \hline
										Drama vs. Comedy & 50,802&215,040,000\\ \hline
									\end{tabular}
								}
							
							\end{subtable}
							
													\end{table}
						
						\subsubsection{Detection Accuracy} We follow the evaluation design of~\cite{Gao2010} and 
	make each baseline methods detect the same percentage of  objects as outliers:
	% as that of the ground-truths. To achieve this, we 
					Sort the outlier scores obtained by the three baseline methods in descending order, and take the top $r$ percent as outliers. Then we use \textbf{precision}, a.k.a. \textbf{true positive rate} as the evaluation metric which is the percentage of correct ones in the set of outliers identified by the algorithm. As in~\cite{Gao2010}, we set the percentages of outlier to be 1\% and 5\%. In the one-class design, precision measures how many members of the outlier class were correctly recognized. We also report some AUC measurements \cite{aggarwal2013}, which aggregate precision values at different percentage cutoffs.\footnote{Our $\mid$ score performs the best also with other metrics such as recall, to a similar degree.} %\textbf{Sarah: I'd love to have the plots of all points here.}

						\paragraph{Likelihood-Based Methods}

						Table~\ref{table:Method Comparison} shows the $\auc$ values for each probabilistic ranking. Our $\mid$ score achieves the top score on each dataset. On the synthetic data, $\mid$ and $|\lr|$ are the only scores with 100\% precision at 1\% and 5\%. This confirms the value of using distances rather than differences. 
%						However, the AUC score shows that $\mid$ retains perfect detection with different percentage cutoffs, but $|\lr|$ ranks some normal objects higher than actual outliers. 
						While it ought to be easy to distinguish the outliers, Table~\ref{table:Metric Comparison} shows that {\em $\mid$  is the only score that achieves perfect detection}, that is AUC = 1.0.

		\begin{table}
																																\caption{Precision of outlier scores in different datasets.
						\label{table:Method Comparison}}
												\resizebox{0.5\textwidth}{!}{	
												\begin{tabular}{|l|c|c|c|c|c|c|c|c|c|}\hline
													Dataset& percentage&\multicolumn{5}{|c|}{Model-based models }&\multicolumn{3}{|c|}{Aggregation-based models}\\
														\hline
																& &\mid&$|\lr|$&\lr&\fd&\loglikelihood&\lof&$\outrank$	&\knn\\
																	\hline
															    \multirow{2}{*}{High-Correlation}& 1\%& \textbf{1.00}& \textbf{1.00}& 0.73& 0.47& 0.91& 0.11& 0.53& 0.48\\
															    &5\%&\textbf{1.00}&\textbf{1.00}&0.85&0.65&0.95&0.22&0.50&0.65\\
														    \hline																				    	 \multirow{2}{*}{Low-Correlation}& 1\%& \textbf{1.00}& \textbf{1.00}& 0.87& 0.14& 0.93& 0.10& 0.00& 0.06\\
												    	    &5\%&\textbf{1.00}&\textbf{1.00}&0.90&0.25&0.95&0.25&0.10&0.14\\
													    	    \hline
																\multirow{2}{*}{Single-Feature}& 1\%& \textbf{1.00}& \textbf{1.00}& 0.39& 0.53& 0.81& 0.46& \textbf{1.00}& 0.51\\
															&5\%&\textbf{1.00}&\textbf{1.00}&0.55&0.62&0.92&0.55&\textbf{1.00}&0.54\\
															\hline    
																\multirow{2}{*}{Striker-Goalie}& 5\%& \textbf{0.57}& 0.27& 0.22& 0.51& 0.36& 0.19& 0.47& 0.42\\
																	&15\%&\textbf{0.63}&0.36&0.31&0.58&0.40&0.32&0.50&0.52\\
																	\hline    
															\multirow{2}{*}{Midfielder-Striker}& 1\%&\textbf{ 0.49}& 0.42& 0.25& 0.41& 0.46& 0.29& 0.44& 0.16\\
															&5\%&\textbf{0.52}&0.48&0.39&0.44&0.50&0.38&0.48&0.35\\
												\hline  																				\multirow{2}{*}{Drama-Comedy}&1\%& \textbf{0.44}& 0.38& 0.39& 0.15& 0.22& 0.29& 0.07& 0.014\\
															&5\%&\textbf{0.47}&0.45&0.44&0.40&0.28&0.36&0.17&0.20\\
																\hline     
																					
												\end{tabular}	}		
																	
						\end{table}

			\begin{table}
				\centering
				\caption{\textit{AUC} of $\mid$ vs. $|LR|$. 
					%For relationships please see text.
					\label{table:Metric Comparison}}
				\resizebox{0.5\textwidth}{!}{
					\begin{tabular}{|l|l|l|l|l|l|l|}\hline
						Score & High-Cor. &Low-Cor.&Single-F.&Striker&Midfielder&Drama\\ \hline
						\mid&1.00&1.00&1.00&0.89&0.66&0.70\\\hline
						$|\lr|$&0.95&0.95&0.89&0.61&0.64&0.65\\\hline
						%	Synthetic Data-High Correlation & \textbf{1.00}&0.95 \\ \hline
						%	Synthetic Data-Low Correlation & \textbf{1.00}&0.95 \\ \hline
						%	Synthetic Data-Single Feature & \textbf{1.00}&0.89 \\ \hline
						%									Real Data-Strikers vs. Goalies & \textbf{0.89} &0.61\\ \hline
						%									Real Data-Midfielders vs. Strikers &\textbf{0.66} &0.64 \\ \hline
						%									Real Data-Drama vs. Comedy & \textbf{0.70} &0.65 \\ \hline
					\end{tabular}}
					
				\end{table}
								\paragraph{Aggregation-Based Methods vs. \mid} 
								Table~\ref{table:Method Comparison} shows the precision values for aggregation-based methods compared to $\mid$. {\em Our $\mid$ score outperforms all aggregation-based methods on all datasets}, except for a tie with $\outrank$(ProClus) on the relatively easy problem of distinguishing strikers from goalies.
%								Comparing tables~\ref{table:Metric Comparison} and~\ref{table:Method Comparison} shows that t
								The performances of aggregation-based methods are most like that of the probabilistic score $\fd$, which does not consider the
								correlation among the features. This finding reflects the fact that aggregation tends to lose information about correlations.
								% \cite{DavidJensen2002}. 
								The aggregation-based methods achieve their highest performance on the Strikers vs. Goalies dataset. In this dataset action count features such as ShotsOnTarget, ShotEfficiency point to strikers and the feature SavesMade points to goalies. Therefore, outliers in this dataset are easy to find by considering features in isolation.
								%\\
								%\\
								%Based on the results shown in table ~\ref{table:Method Comparison}, the easiest dataset is Single Features, as most of the methods are successful in detecting outliers in that dataset. The low correlation dataset seems to be the most challenging one. 
								%The results of $\knn$ seem to be closest to $\mid$ in most datasets. 
								
								%The $\roc$ curve in Figure~\ref{fig:ROC} shows the performance of $\mid$ and $\knn$ which indicates that $\mid$ always reaches a higher true positive rate earlier that $\knn$.
								%\textbf{Sarah: both graph and table?}
								%Show the running time in a table. Explain that thats one of the weakness of $\mid$ compare to the others. But since the its performance is much higher than the compatitors it is worth the waiting!
								
								%
								%
								%
								%Strikers vs. goalies dataset is very similar to synthetic dataset, single feature in a sense that outlier and normal classes in both datasets contain single features that have distinct and very different distributions independent of the other attributes. In Strikers vs. Goalies attributes such as ShotsOnTarget or ShotEfficiency are valid only if the object is a member of Striker class and SavesMade only belongs to Goalie Class. Therefore, outliers in this dataset are also very easy to find even by applying the aggregation-based methods.  

								%\begin{figure*}
								%\centering
								%\resizebox{0.9\textwidth}{!}{
								
								%\subfigure{
								%  \includegraphics[height=70mm, width=70mm] {figures/TPR-Synthetic.pdf}
								%}
								%\subfigure{
								%  \includegraphics[height=70mm, width=70mm] {figures/TPR-All.pdf}
								
								% }
								% }
								
								%\caption{Comparison of Object Outlier Metrics}
								%\label{fig:synthetic}
								%\end{figure*}
								
								\subsubsection{Case Studies} For a case study, we examine three top outliers as ranked by $\mid$, shown in Table~\ref{table:CaseStudy}. 
								The aim of the case study is to provide a qualitative sense of the outliers indicated by the scores. Also, we illustrate how the BN representation leads to an interpretable ranking. 
							Specifically, we employ a {\em feature-wise decomposition} of the score combined with a {\em drill down} analysis: 
							
\begin{enumerate}
\item Find the node $\feature_{i}$ that has the highest $\mid_{i}$ divergence score for the outlier object. 
\item Find the parent-child combination that contributes the most to the $\mid_{i}$ score for that node.
\item Decompose the $\mid$ score for the parent-child combination into feature and mutual information component. 
\end{enumerate}
	
								We present strong associations---indicated by the $\mid$'s mutual information component---in the intuitive format of association rules.
								
								%\begin{figure}
								%\centering
								%   \includegraphics[width=0.5\textwidth] {figures/ROCKNN.pdf}
								% \caption{Detection Accuracy of $\mid$ vs. $\knn$
								% \label{fig:ROC}}
								%\end{figure}
								%\vspace{-5mm}
								\paragraph{Strikers vs. Goalies} 
								%The Mutual Information Score $\mid$ separates goalies from Strikers better compared to the other methods.  
								%
								In real-world data, a rare object may be a {\em within-class outlier}, i.e., highly anomalous even within its class. In an unsupervised setting without class labels, we do not expect an outlier score to distinguish such an in-class outlier from outliers outside the class. 
%								This is the reason why in real-world data, we do not expect an outlier detection score to distinguish the normal class objects perfectly from objects outside the class. 
								An example is the striker Edin Dzeko. He is a highly anomalous striker who obtains 
								the top $\mid$ divergence score among both strikers and goalies. His $\mid$ score is highest for the Dribble Efficiency feature. The highest $\mid$ score for that feature occurs when Dribble Efficiency is low, and its parents have the following values: Shot Efficiency high, Tackle Efficiency medium. Looking at the single feature divergence, 
								%Decomposing this $\mid$ score into feature divergence and joint information divergence, 
								we see that Edin Dzeko is indeed an outlier in the Dribble Efficiency subspace: His dribble efficiency is low in 16\% of his matches, whereas a randomly selected striker has low dribble efficiency in 50\% of their matches. Thus, Edin Dzeko is an unusually good dribbler. Looking at the mutual information component of $\mid$, i.e., the parent-child correlations, for Edin Dzeko the confidence of the rule 
								$$\it{ShotEff} = \it{high}, \it{TackleEff} = \it{medium}\rightarrow \it{DribbleEff} = \it{low}$$ is 50\%, whereas in the general striker class it is $38\%$.
								%The $\eld$ divergence also ranks Edin Dzeko as unusual. But because it allows feature and joint information divergence to cancel, his rank is somewhat lower. The likelihood metric does not recognize him as unusual at all. 
								
%								The next two outliers according to $\mid$ are goalies Paul Robinson and Michel Vorm. Their rank is based only on feature divergence, with zero mutual information distinction. The maximum feature divergence is obtained by the $\it{SavesMade}$ feature. This makes intuitive sense since strikers basically never make saves. 
								%In other words, feature divergence with respect to $\it{SavesMade}$ is a good way to distinguish goalies from strikers. 
								%
								%The $\eld$ divergence also ranks Paul Robinson and Michel Vorm as clear goalies.  The likelihood metric does not recognize Paul Robinson as unusual at all. 
								%\vspace{-5mm}
								\paragraph{Midfielders vs. Strikers} 
								%The  $\mid$ metric separates midfielders from strikers better compared than the other methods.  
								%The single feature divergence does not discriminate these two classes of objects. Intuitively, this is because strikers and midfielders are generally similar with respect to single features.  
								%The distance metrics have a better TOR rate than the averaging metrics. 
								
							%	The decomposition analysis for the top three $\mid$ outliers proceeds as follows. 
								For the single feature score, Robin van Persie is recognized as a clear striker because of the $\it{ShotsOnTarget}$ feature. It makes sense that strikers shoot on target more often than midfielders. Robin van Persie  achieves a high number of shots on targets in $34\%$ of his matches, compared to $3\%$ for a random midfielder. The mutual information component shows that he also exhibits  unusual correlations. For example, 
								the confidence of the rule
								$$\it{ShotEff} = \it{high}, \it{TimePlayed} = \it{high} \rightarrow \it{ShotsOnTarget} = \it{high}$$
								is 70\% for van Persie, whereas for strikers overall it is 52\%.
								%Both the $\eld$ metric and the $\lnlikelihood$ metric recognize Van Persie as a striker. 
								
%								Wayne Rooney is recognized as a striker for similar reasons, but less clearly because he achieves a high number of shots on target less frequently. 
The most anomalous midfielder is Scott Sinclair. His most unusual feature is $\it{DribbleEfficiency}$: For feature divergence, he achieves a high dribble efficiency $50\%$ of the time, compared to a random midfielder with $30\%$. 
								%The $\jid$ divergence shows that he also exhibits unusual correlations for DribbleEfficiency.
								%The $\eld$ divergence too ranks Scott Sinclair as an unusual midfielder, whereas the likelihood method places him in the middle of his class. 
								%\vspace{-5mm}
								\paragraph{Drama vs. Comedy} 
								%As with the other datasets, the  $\mid$ metric separates normal objects  from the contrast class better than the other methods.   
								The top outlier rank is assigned to the within-class outlier $\it{Brave Heart}$. Its most  unusual feature is  $\it{ActorQuality}$: In a random drama movie,  $42\%$ of actors have the highest quality level 4, whereas for $\it{Brave Heart}$ $93\%$ of actors achieve the highest quality level. 
								%The $\eld$ divergence also ranks $\it{Brave Heart}$ as an unusual drama, whereas the likelihood method places it in the middle of its class. 
								
								%based on user assigned ranking. 
								The  $\mid$ score identifies the comedies  $\it{Blues Brothers}$ and $\it{Austin Powers}$ as the top out-of-class outliers. 
							%	The main contributor to these rankings is the $\it{Cast\_Position}$ feature. 
								In a random drama movie,  $49\%$ of actors have casting position 3, whereas for $\it{Austin Powers}$ $78\%$ of actors have this casting position, and for $\it{Blues Brothers}$ $88\%$ of actors do. 
%								These three movies also show unusual correlations for this feature with high divergence in the mutual information component (not shown in Table~\ref{table:CaseStudy}).
								\begin{table*}
									\centering
									\caption{Case study for the top outliers returned by the log-likelihood distance score \mid
										\label{table:CaseStudy}}
									\resizebox{0.8\textwidth}{!}{
										\begin{tabular}{|l|l|l|l|l|l|l|l|} \hline
											\multicolumn{8}{|c|}{Strikers (Normal) vs. Goalies (Outlier)}\\
											\hline
											PlayerName&Position&$\mid$ Rank&$\mid$ Max Node&$\mid$ Node Score&$\fd$ Max feature Value& Object Probability& Class Probability \\ \hline
											Edin Dzeko&Striker&1&DribbleEfficiency&83.84&DE=low&0.16&0.5 \\ \hline
											Paul Robinson&Goalie&2&SavesMade&49.4&SM=Medium&0.3&0.04\\ \hline
											Michel Vorm&Goalie&3&SavesMade&85.9&SM=Medium&0.37&0.04\\ \hline
											\multicolumn{8}{|c|}{Midfielders (Normal) vs. Strikers (Outlier)}\\
											\hline
											PlayerName&Position&$\mid$ Rank&$\mid$ Max Node&$\mid$ Node Score&$\fd$ Max feature Value& Object Probability& Class Probability \\ \hline
											Robin Van Persie& Striker&1&ShotsOnTarget&153.18&ST=high&0.34&0.03 \\ \hline
											Wayne Rooney& Striker&2&ShotsOnTarget&113.14&ST=high&0.26&0.03\\ \hline
											Scott Sinclair&Midfielder&6&DribbleEfficiency&71.9&DE=high&0.5&0.3\\ \hline
											\multicolumn{8}{|c|}{Drama (Normal) vs. Comedy (Outlier)}\\
											\hline
											MovieTitle&Genre&$\mid$ Rank&$\mid$ Max Node&$\mid$ Node Score& $\fd$ Max feature Value& Object Probability& Class Probability \\ \hline
											Brave Heart&Drama&1&ActorQuality&89995.4&a\_quality=4&0.93&0.42\\ \hline
											Austin Powers&Comedy&2&Cast\_Position&61021.28&Cast\_Num=3&0.78&0.49\\ \hline
											Blue Brothers&Comedy&3&Cast\_Position&24432.21&Cast\_num=3&0.88&0.49\\ \hline
										\end{tabular} 
									}
								\end{table*}

\section{Conclusion} We presented a new approach for applying Bayes nets to object-relational outlier detection, a challenging and practically important topic for machine learning. This approach follows the general framework of Exceptional Model Mining~\cite{Duivesteijn2016}, and applies it to multi-relational data. The key idea is to learn one set of parameter values that represent class-level associations, another set to represent object-level associations, and compare how well each parametrization fits the relational data that characterize the target object. The classic metric for comparing two parametrized models is their log-likelihood ratio; we refined this concept to define  a new relational log-likelihood distance metric via two transformations:  (1) a mutual information decomposition, and (2) replacing log-likelihood differences by log-likelihood distances. This metric combines a single feature component, where features are treated as independent, with a correlation component that measures the deviation in the features' mutual information.

In experiments on three synthetic and three real-world outlier sets, the log-likelihood distance achieved the best detection accuracy. The alternative of converting the structured data to a flat data matrix via aggregation had a negative impact. %on outlier detection. 
Case studies showed that the log-distance score leads to easily interpreted rankings.
%
%Overall, our new log-likelihood distance metric provides a promising new approach for applying machine learning techniques to outlier detection for object-relational data, a challenging and practically important topic. 

								There are several avenues for future work.  (i) A limitation of our current approach is that it ranks potential outliers, but does not set a threshold for a binary identification of outlier vs. non-outlier. (ii) Our divergence uses expected L1-distance for interpretability, but other distance scores like L2 could be investigated as well. (iii) Extending the expected L1-distance for continuous features is a useful addition. 
								%, and may facilitate combining our object-oriented approach with dimensional hierarchies in an OLAP data cube.
								%Distribution distances other than KLD could be evaluated for outlier detection (e.g., variation distance). However, the KLD variants have special advantages: their asymetry reflects the asymetry between object and class. Also, in a Bayesian network representation of the joint distributions, they decompose into a node-wise sum for easy computation and interpretation. 
								%A promising project for future work would be to combine the object data model and the multi-dimensional model to combine our object outlier method with OLAP-based methods. For example, one could add numeric measures and aggregate attributes to the object model. Another view: convert the object-oriented data to OLAP data, then use their outlier detection.

								In sum, outlier metrics based on model likelihoods are a new type of structured outlier score for object-relational data.  Our evaluation indicates that this model-based score provides informative, interpretable, and accurate rankings of objects as potential outliers. 
								
				\section*{Acknowledgement}
This work was supported by a Discovery Grant from the National Sciences and Engineering Council of Canada. We are indebted to Peter Flach for referring us to the EMM framework.

								\bibliographystyle{abbrv}
								\bibliography{Bibliography-File,master}
							\end{document}

%% file: preamble-stuff.tex
\usepackage{amsmath}
\usepackage{amsfonts}
\usepackage{amssymb}
\usepackage{graphicx}
\usepackage{url}
\usepackage{epstopdf}
\usepackage{subfigure}
\usepackage{fancyhdr}
\graphicspath{{../}{figures/}}

\renewcommand{\P}{P}

\renewcommand{\v}{v}

\newcommand{\loglikelihood}{\it{LOG}}
\newcommand{\lof}{\it{LOF}}

\newcommand{\lr}{\it{LR}}

\renewcommand{\mid}{\it{ELD}}

\newcommand{\outrank}{\it{OutRank}}
\newcommand{\knn}{\it{KNNOutlier}}
\newcommand{\auc}{\it{Precision}}

\newcommand{\fd}{\it{FD}}
\newcommand{\parameter}{\theta}
\newcommand{\parameters}{\bs{\parameter}}

\newcommand{\nodevalue}{\v}
\newcommand{\parents}{\mathbf{pa}}

\newcommand{\states}{r} % number of states of a variable
\newcommand{\functor}{f}

\newcommand{\population}{\mathcal{P}}

\newcommand{\term}{\sigma}

\newcommand{\grounds}{\#}

\newcommand{\class}{c} % the class attribute
\newcommand{\object}{o}
\newcommand{\Class}{C}

\newcommand{\model}{B}

\newcommand{\feature}{V} % feature or desc attribute of object or link
\newcommand{\Features}{\bs{V}}
\newcommand{\M}{M}

\newcommand{\T}{T} % for tables generically

\newcommand{\D}{\mathcal{D}} % for database instance
\newcommand{\team}{\it{T}}
\newcommand{\player}{\it{P}}
\newcommand{\match}{\it{M}}

\newcommand{\director}{\it{Director}}
\newcommand{\movie}{\it{Movie}}
\newcommand{\user}{\it{User}}

\newcommand{\actor}{\mathit{Actor}}

\newcommand{\true}{\mathit{T}}
\newcommand{\false}{\mathit{F}}
\def\set#1{\mathbf{#1}}
\def\bs#1{\boldsymbol{#1}}

%% file: starai-2018.bbl
\begin{thebibliography}{10}

\bibitem{Elke2013}
E.~Achtert, H.~Kriegel, E.~Schubert, and A.~Zimek.
\newblock Interactive data mining with 3d-parallel coordinate trees.
\newblock In {\em Proceedings of the 2013 ACM SIGMOD}, New York, NY, USA, 2013.

\bibitem{aggarwal2013}
C.~Aggarwal.
\newblock {\em Outlier Analysis}.
\newblock Springer New York, 2013.

\bibitem{Akoglu2015}
L.~Akoglu, H.~Tong, and D.~Koutra.
\newblock Graph based anomaly detection and description: a survey.
\newblock {\em Data Mining and Knowledge Discovery}, 29(3):626--688, 2015.

\bibitem{Angiulli2007}
F.~Angiulli, G.~Greco, and L.~Palopoli.
\newblock Outlier detection by logic programming.
\newblock {\em ACM Transactions on Computer Logic}, 2004.

\bibitem{Breunig2000}
M.~Breunig, H.-P. Kriegel, R.~T. Ng, and J.~Sander.
\newblock Lof: Identifying density-based local outliers.
\newblock In {\em Proceedings of ACM SIGMOD}, 2000.

\bibitem{Cansado2008}
A.~Cansado and A.~Soto.
\newblock Unsupervised anomaly detection in large databases using {Bayes Nets}.
\newblock {\em Appllied Artificial Intelligence}, 2008.

\bibitem{de2016bayeswipe}
S.~De, Y.~Hu, V.~V. Meduri, Y.~Chen, and S.~Kambhampati.
\newblock Bayeswipe: A scalable probabilistic framework for improving data
  quality.
\newblock {\em Journal of Data and Information Quality (JDIQ)}, 8(1):5, 2016.

\bibitem{Campos2006}
L.~de~Campos.
\newblock A scoring function for learning {B}ayes nets based on mutual
  information and conditional independence tests.
\newblock {\em Journal of Machine learning Research}, 2006.

\bibitem{Domingos2009}
P.~Domingos and D.~Lowd.
\newblock {\em Markov Logic: An Interface Layer for Artificial Intelligence}.
\newblock Morgan and Claypool Publishers, 2009.

\bibitem{Duivesteijn2016}
W.~Duivesteijn, A.~J. Feelders, and A.~Knobbe.
\newblock Exceptional model mining.
\newblock {\em Data Mining and Knowledge Discovery}, 30(1):47--98, 2016.

\bibitem{Flach1999a}
P.~A. Flach.
\newblock Knowledge representation for inductive learning.
\newblock In {\em Symbolic and Quantitative Approaches to Reasoning and
  Uncertainty}, pages 160--167. Springer, 1999.

\bibitem{Gao2010}
J.~Gao, F.~Liang, W.~Fan, Y.~Wang, and J.~Han.
\newblock On community outliers and their detection in information network.
\newblock In {\em Proceedings of ACM SIGKDD}, 2010.

\bibitem{SRL2007}
L.~Getoor and B.~Taskar.
\newblock {\em Introduction to statistical relational learning}.
\newblock MIT Press, 2007.

\bibitem{IMDb-original}
{Internet Movie Database}.
\newblock Internet movie database.
\newblock [Online]. Available: URL = \url{http://www.imdb.com/}.

\bibitem{bib:jbnsite}
H.~Khosravi, T.~Man, J.~Hu, E.~Gao, and O.~Schulte.
\newblock Learn and join algorithm code.
\newblock [Online]. Available: URL = \url{http://www.cs.sfu.ca/~oschulte/jbn/}.

\bibitem{Kimmig2014}
A.~Kimmig, L.~Mihalkova, and L.~Getoor.
\newblock Lifted graphical models: a survey.
\newblock {\em Computing Research Repository}, 2014.

\bibitem{Koh2008}
J.~L. Koh, M.~L. Lee, W.~Hsu, and W.~T. Ang.
\newblock Correlation-based attribute outlier detection in {XML}.
\newblock In {\em Proceedings of ICDE. IEEE 24th}, 2008.

\bibitem{Koller1997}
D.~Koller and A.~Pfeffer.
\newblock Object-oriented {B}ayes nets.
\newblock In {\em Proceedings of UAI}, 1997.

\bibitem{Maervoet2012}
J.~Maervoet, C.~Vens, G.~Vanden~Berghe, H.~Blockeel, and P.~De~Causmaecker.
\newblock Outlier detection in relational data: A case study.
\newblock {\em Expert System Applications}, 2012.

\bibitem{Mayfield2010}
C.~Mayfield, J.~Neville, and S.~Prabhakar.
\newblock Eracer: a database approach for statistical inference and data
  cleaning.
\newblock In {\em Proceedings of the 2010 ACM SIGMOD International Conference
  on Management of data}, pages 75--86. ACM, 2010.

\bibitem{opta-original}
{MCFC Analytics}.
\newblock The premier league dataset.
\newblock [Online]. Available: URL =
  \url{http://www.mcfc.co.uk/Home/MCFCAnalytics}.

\bibitem{Muller2012}
E.~Muller, I.~Assent, P.~Iglesias, Y.~Mulle, and K.~Bohm.
\newblock Outlier ranking via subspace analysis in multiple views of the data.
\newblock In {\em Proceedings of ICDM}, 2012.

\bibitem{Novak2009}
P.~K. Novak, G.~I. Webb, and S.~Wrobel.
\newblock Supervised descriptive rule discovery: A unifying survey of contrast
  set, emerging pattern and subgroup mining.
\newblock {\em Journal of Machine Learning Research}, 2009.

\bibitem{Pearl1988}
J.~Pearl.
\newblock {\em Probabilistic Reasoning in Intelligent Systems}.
\newblock Morgan Kaufmann, 1988.

\bibitem{Peralta2007}
V.~Peralta.
\newblock {Extraction and Integration of MovieLens and IMDb}.
\newblock Technical report, APDM project, 2007.

\bibitem{Poole2003}
D.~Poole.
\newblock First-order probabilistic inference.
\newblock In {\em Proceedings of IJCAI}, 2003.

\bibitem{Ramaswamy2000}
S.~Ramaswamy, R.~Rastogi, and K.~Shim.
\newblock Efficient algorithms for mining outliers from large data sets.
\newblock {\em SIGMOD}, 2000.

\bibitem{url}
F.~Riahi and O.~Schulte.
\newblock {Codes and Datasets. [Online]. Available:}.
\newblock \url{ftp://ftp.fas.sfu.ca/pub/cs/oschulte/CodesAndDatasets/}, 2015.

\bibitem{Riahi2015}
F.~Riahi and O.~Schulte.
\newblock Model-based outlier detection for object-relational data.
\newblock In {\em 2015 IEEE Symposium Series on Computational Intelligence},
  pages 1590--1598. IEEE, 2015.

\bibitem{Sarawagi1998}
S.~Sarawagi, R.~Agrawal, and N.~Megiddo.
\newblock Discovery-driven exploration of {OLAP} data cubes.
\newblock In {\em Proceedings of International Conference on Extending Database
  Technology}. Springer-Verlag, 1998.

\bibitem{Schulte2011}
O.~Schulte.
\newblock A tractable pseudo-likelihood function for {Bayes} nets applied to
  relational data.
\newblock In {\em Proceedings of SIAM SDM}, 2011.

\bibitem{Schulte2012}
O.~Schulte and H.~Khosravi.
\newblock Learning graphical models for relational data via lattice search.
\newblock {\em Journal of Machine Learning}, 2012.

\bibitem{Schulte2012a}
O.~Schulte, H.~Khosravi, and T.~Man.
\newblock Learning directed relational models with recursive dependencies.
\newblock {\em Machine Learning}, 89:299--316, 2012.

\bibitem{Tang2013}
G.~Tang, J.~Bailey, J.~Pei, and G.~Dong.
\newblock Mining multidimensional contextual outliers from categorical
  relational data.
\newblock In {\em Proceedings of SSDBM}, 2013.

\bibitem{Tuffery2011}
S.~Tuffery.
\newblock {\em Data Mining and Statistics for Decision Making}.
\newblock Wiley Series in Computational Statistics, 2011.

\end{thebibliography}
